\documentclass[10pt,twocolumn,letterpaper]{article}

\usepackage{iccv}
\usepackage{times}
\usepackage{epsfig}
\usepackage{graphicx}
\usepackage{amsmath}
\usepackage{amssymb}
\usepackage{booktabs}
\usepackage{multicol}
\usepackage{multirow}
\usepackage{makecell}
\usepackage{amsmath}
\usepackage{amsthm,enumitem}
\usepackage[accsupp]{axessibility}  

\usepackage[pagebackref=true,breaklinks=true,letterpaper=true,colorlinks,bookmarks=false]{hyperref}

\usepackage[capitalize]{cleveref}
\crefname{section}{Sec.}{Secs.}
\Crefname{section}{Section}{Sections}
\Crefname{table}{Table}{Tables}
\crefname{table}{Tab.}{Tabs.}

\newcommand{\MyMethod}{\emph{S-VolSDF}}
\newcommand{\myP}{{\mathbf{P}}}
\newcommand{\myPprime}{{\mathbf{P^{\prime}}}}
\newcommand{\myx}{{\mathbf{x}}}
\newcommand{\myw}{{\mathbf{w}}}

\iccvfinalcopy 

\def\iccvPaperID{9034} 

\ificcvfinal\fi

\begin{document}

\setlength{\abovedisplayskip}{3pt}
\setlength{\belowdisplayskip}{3pt}

\title{S-VolSDF: Sparse Multi-View Stereo Regularization of Neural Implicit Surfaces}


\author{
Haoyu Wu \qquad Alexandros Graikos \qquad Dimitris Samaras\\
Stony Brook University\\
{\tt\small \{haoyuwu,agraikos,samaras\}@cs.stonybrook.edu}
}

\maketitle
\ificcvfinal\thispagestyle{empty}\fi

\begin{abstract}
Neural rendering of implicit surfaces performs well in 3D vision applications. However, it requires dense input views as supervision. When only sparse input images are available, output quality drops significantly due to the shape-radiance ambiguity problem. We note that this ambiguity can be constrained when a 3D point is visible in multiple views, as is the case in multi-view stereo (MVS). We thus propose to regularize neural rendering optimization with an MVS solution. The use of an MVS probability volume and a generalized cross entropy loss leads to a noise-tolerant optimization process. In addition, neural rendering provides global consistency constraints that guide the MVS depth hypothesis sampling and thus improves MVS performance. Given only three sparse input views, experiments show that our method not only outperforms generic neural rendering models by a large margin but also significantly increases the reconstruction quality of MVS models.
\end{abstract}


\begin{figure}
\begin{center}
\includegraphics[width=0.47\textwidth]{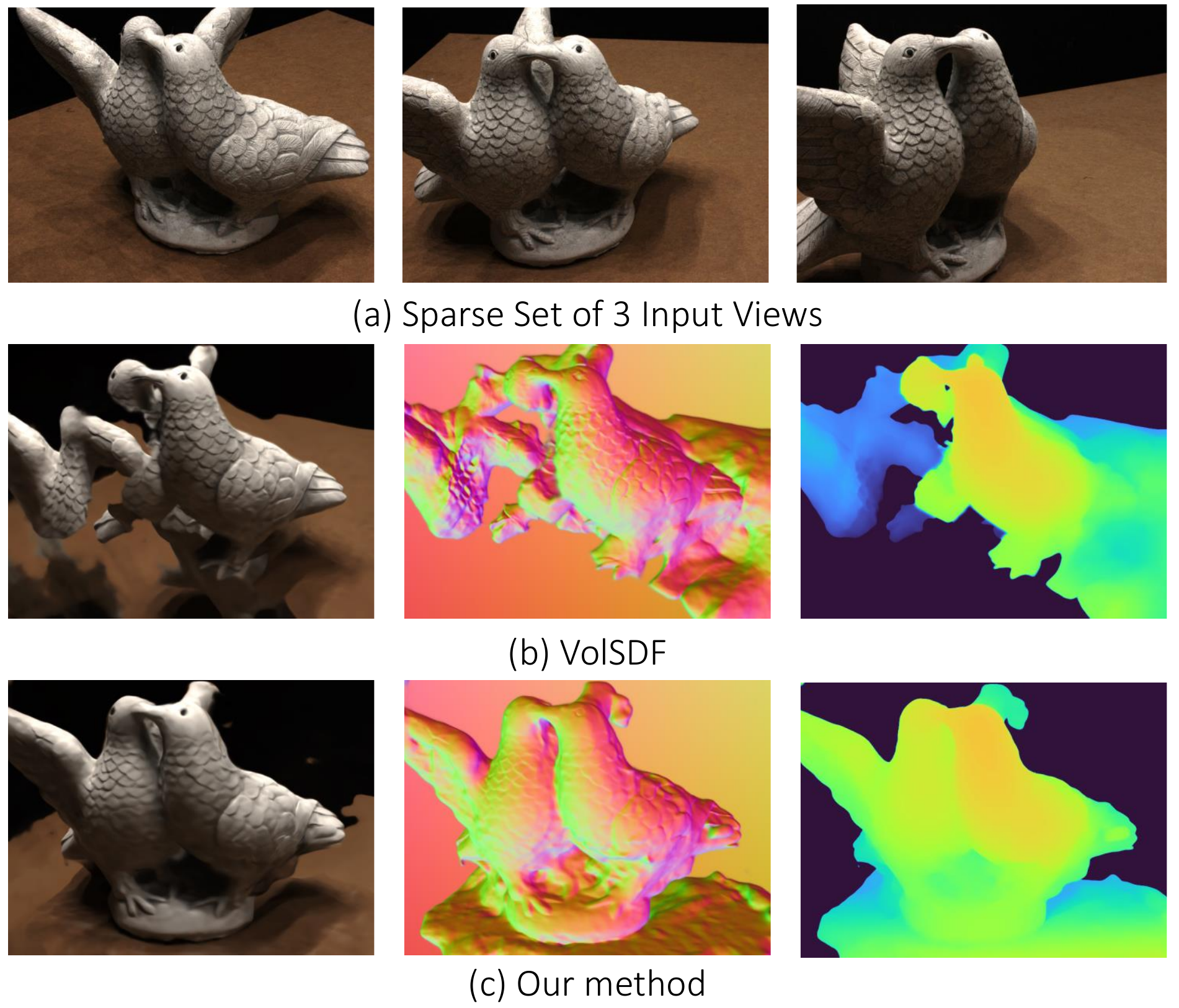}
\end{center}
   \caption{Shape-Radiance Ambiguity. In the last two rows, we compare the novel view synthesis results from VolSDF \cite{yariv2021volume} and our model: RGB renderings (left), predicted normal maps (middle), and expected depth maps (right).}
\label{fig:nerf_demo}
\end{figure}

\section{Introduction}
\label{sec:Introduction} 

Neural surface reconstruction techniques, coupled with coordinate-based neural network models, have become increasingly popular in the field of 3D vision \cite{yariv2021volume, wang2021neus, yariv2020multiview}. Although these methods perform very well, they require dense input views as supervision. This is limiting for many real-world applications where sparse input images are the only source of information, such as robotics, augmented reality, autonomous driving, and scene reconstruction in-the-wild. As shown in Fig. \ref{fig:nerf_demo}, the reconstruction quality of a scene using VolSDF \cite{yariv2021volume} (a state-of-the-art technique) drops significantly when only 3 views are used. This is due to the {\em shape-radiance ambiguity} problem \cite{zhang2020nerf++}. 


The {\em shape-radiance ambiguity} \cite{zhang2020nerf++} means that there is a high probability an incorrect geometry reconstruction satisfies the photometric constraint when it is visible from a single view only, as is in the case of sparse views. In that scenario, the photometric loss alone can not guide the model toward a correct solution. To regularize this, we need to constrain surface points to be visible from multiple views, hence, we need correspondences, as in multi-view stereo (MVS) \cite{chen2019point,luo2019p, xu2020learning, yao2018mvsnet, yao2019recurrent, cheng2020deep, yu2020fast, gu2020cascade, yang2020cost, zhang2020visibility, wei2021aa, yan2020dense, wang2022mvster, zhang2021long, zhu2021multi, ding2022transmvsnet}. Thus, we propose to guide neural rendering optimization with information from MVS. The challenge is how to effectively incorporate the noisy MVS predictions into the neural rendering pipeline.

Many modern MVS methods \cite{yao2018mvsnet, gu2020cascade, ding2022transmvsnet, cheng2020deep} integrate the evidence for each possible 3D point into a probability volume and regress depth from it. In order to avoid possible errors in MVS 3D point reconstruction, we do not use point estimates, but the whole probability volume. We also note that the rendering weights in neural rendering methods and the probability volume in MVS actually have the same meaning: the probability that a point at a particular location is visible by multiple views. Based on recent MVS literature \cite{peng2022rethinking}, we can think of all possible 3D points on a ray as interior or exterior to the object (i.e. a binary classification problem). Thus, we can treat the MVS probability volume as a set of noisy labels for the rendering weights (i.e. occupancy values). Posing neural rendering as a classification problem allows the use of cross entropy loss to optimize neural rendering methods. However, as shown by the classification literature \cite{zhang2018generalized, song2022learning}, the cross entropy loss is sensitive to noisy labels. Instead, we adopt a generalized cross entropy loss \cite{zhang2018generalized} to reduce the penalty on false positive MVS predictions and thus increase the optimization's tolerance to noise.

In order to produce our final geometry, we want to take advantage of global consistency constraints including photometric consistency and surface smoothness imposed by neural rendering. Thus, we propose to incorporate neural surface reconstruction into coarse-to-fine MVS models. Specifically, we use the coarse stage MVS predictions to regularize neural surface optimization. Then, we use the rendered depth maps to guide the next stage's depth hypothesis sampling in MVS. Moreover, neural surface optimization only requires 10-15 minutes in current hardware to obtain good results because of strong geometry cues from MVS. As a result, we obtain much better surface reconstruction than either MVS or neural rendering alone, at a relatively fast speed.

In this paper, we propose \MyMethod{}, a novel approach that leverages multi-view stereo priors to optimize neural surface reconstruction with sparse input views. Our main contributions are as follows:
\begin{itemize}
\item We propose a simple but effective noise-tolerant cost function that combines multi-view stereo with neural volumetric surface reconstruction methods, so their optimization is regularized by the probability volumes of MVS methods.
\item We integrate neural surface reconstruction into multiple coarse-to-fine MVS models. Our method consistently improves depth estimation for better MVS performance at a faster speed.
\item We evaluate our method on surface reconstruction and novel view synthesis on the DTU \cite{aanaes2016large} and BlendedMVS \cite{yao2020blendedmvs} datasets. Our reconstruction is significantly better than both neural rendering and MVS models.
\end{itemize}

\section{Related Work}

\subsection{Multi-View Stereo (MVS)}
\label{sec: Multi-View Stereo}

Traditional multi-view stereo uses representations such as depth maps, point clouds, and volumetric representations \cite{furukawa2015multi}. Depth map based methods \cite{campbell2008using,galliani2015massively,tola2012efficient,schonberger2016pixelwise,xu2019multi} typically rely on a reference image and additional nearby source images for depth estimation. Point cloud based methods \cite{furukawa2009accurate, lhuillier2005quasi, locher2016progressive} attempt to optimize a collection of patches that best describe a 3D scene. Volumetric methods \cite{kendall2017end, kar2017learning, kutulakos2000theory, seitz1999photorealistic, curless1996volumetric, vogiatzis2005multi, hernandez2007probabilistic, sinha2007multi, furukawa2009reconstructing, zach2007globally} often aggregate information into a global representation such as a volume or mesh. 

Deep-learning MVS methods \cite{chen2019point,luo2019p, xu2020learning, yao2018mvsnet, yao2019recurrent, cheng2020deep, yu2020fast, gu2020cascade, yang2020cost, zhang2020visibility, wei2021aa, yan2020dense, wang2022mvster, zhang2021long, zhu2021multi, ding2022transmvsnet} typically use depth maps as 3D representations and follow the steps below: i) they use a differentiable homography to aggregate features from nearby views and build the cost volume, ii) they use a 3D CNN to regularize the cost volume and regress the depth and finally, iii) by applying a \emph{softmax} function, they obtain a probability volume from the cost volume. A winner-takes-all technique is often used to determine the depth. Cascade cost volumes \cite{cheng2020deep, gu2020cascade, yang2020cost, zhang2020visibility, ding2022transmvsnet} and recurrent cost volume regularization \cite{wei2021aa, yan2020dense, yao2019recurrent} further reduce memory consumption. The cascade cost volume is constructed in a coarse-to-fine manner that first regresses a coarse depth in low resolution and then predicts finer depth values in higher resolution based on the depth range inferred from the coarse result.

MVS explicitly forces surface points to be visible from multiple views. This property prevents degenerate geometry in the case of sparse input views. However, the correspondence problem is often hard to solve, which introduces significant noise in the predicted geometry. Furthermore, the use of the \emph{argmax} operation (winner-takes-all) removes potentially correct predictions in the MVS probability volume and introduces further noise. Thus, we propose to directly use information from the probability volume instead of the noise-prone MVS point estimates.

\subsection{Neural Volumetric Representations}
\label{sec: Neural Volumetric Representations}

Neural volumetric representations are popular in 3D reconstruction \cite{jiang2020sdfdiff, yariv2020multiview, niemeyer2020differentiable, kellnhofer2021neural, liu2020dist, wang2021neus, yariv2021volume, oechsle2021unisurf, zhang2021learning, darmon2022improving} and novel view synthesis \cite{sitzmann2019scene, lombardi2019neural, mildenhall2020nerf, saito2019pifu, trevithick2020grf, barron2021mip, martin2021nerf, park2019deepsdf, yariv2020multiview}. NeRF \cite{mildenhall2020nerf}, the most well-known method, is based on the volume rendering equation \cite{max1995optical, kajiya1984ray} and stores 3D information inside a neural network in the form of a compact Multi-layer Perceptron (MLP). Due to the expressive power of the neural network, it is able to model high-quality details and reconstruct complex 3D structures with a relatively small storage cost. VolSDF builds on NeRF with improved volumetric rendering of implicit surfaces. However, as shown in Fig.~\ref{fig:nerf_demo}, in the case of sparse input views, the quality of the VolSDF reconstruction drops significantly because of the radiance-ambiguity problem described in \cref{sec:Introduction}. 

Regularization-based approaches are simple, but efficient ways to mitigate this problem, using priors such as smoothness \cite{niemeyer2022regnerf, oechsle2021unisurf}, cross-view semantic similarity constraint \cite{jain2021putting}, normal priors \cite{wang2022neuris}, and depth priors \cite{roessle2022dense, wei2021nerfingmvs}. DS-NeRF \cite{deng2022depth} utilizes estimated depth from structure-from-motion \cite{schonberger2016structure}. MonoSDF \cite{yu2022monosdf} and SparseNeRF \cite{wang2023sparsenerf} utilize monocular depth estimation. Monocular depth estimation is often not accurate, only roughly approximating shapes, and may lead to sub-optimal results. Sensor depth \cite{dey2022mip, kirmayr2022dgdnerf} and MVS \cite{zhang2021learning, zuo2022view} have also been adopted to regularize the training of neural rendering models. Although MVS is a strong prior in general, it can be unreliable when the MVS prediction is noisy with sparse input views.


A different approach is to increase the generalization ability of neural rendering by utilizing priors derived from a larger model trained on multi-view image datasets \cite{long2022sparseneus, chen2021mvsnerf, wang2021ibrnet, yu2021pixelnerf, liu2022neural, chibane2021stereo, liu2022neural, suhail2022generalizable}. PixelNeRF \cite{yu2021pixelnerf} is conditioned on features extracted by a CNN. MVSNeRF \cite{chen2021mvsnerf} forms a neural volume from the cost volume obtained by warping image features, and is conditioned on this neural volume. IBRNet \cite{wang2021ibrnet} aggregates features from nearby views to infer geometry and adopts an image-based rendering approach. GeoNeRF \cite{johari2022geonerf} utilizes a cascaded cost volume and an attention-based technique to aggregate information from different views. SparseNeuS \cite{long2022sparseneus} proposes cascaded geometry reasoning and consistency-aware fine-tuning. These methods considerably improve reconstruction, but our experiments show that their results still suffer from entanglement of texture with geometry, and inconsistencies between views.

Our method differs from generic neural rendering methods like MVSNeRF and GeoNeRF in that we explicitly utilize the MVS prior through noise-tolerant test-time optimization. In contrast, generic methods implicitly utilize the MVS prior by conditioning the rendering MLP on features derived from the cost volume, which may not work well in challenging sparse-input scenarios. In \cref{sec: Comparisons}, we show our approach outperforms generic methods to effectively and reliably disentangle texture and geometry.


\section{Method}                               
\label{sec: Method}

\begin{figure}[!htb]
\begin{center}
\centerline{\includegraphics[width=0.5\textwidth]{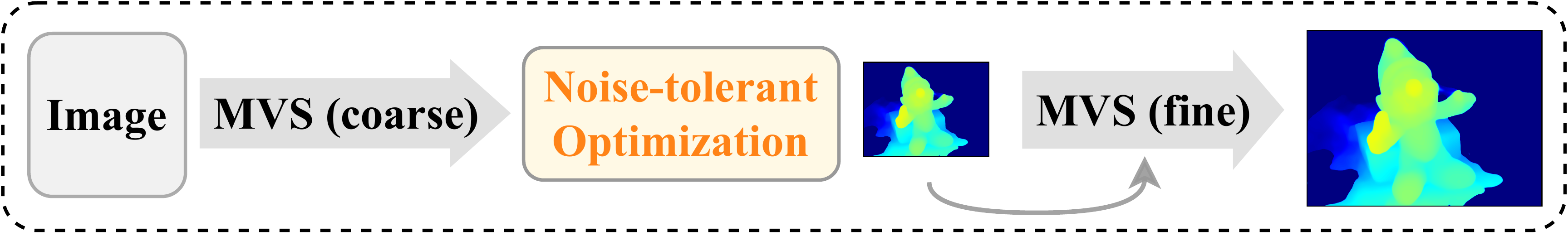}}
\end{center}
\caption{Our proposed method improves the quality of depth maps obtained from the coarse stage multi-view stereo (MVS) by introducing noise-tolerant optimization techniques. The resulting depth maps then guide depth hypothesis sampling in the finer stage MVS, leading to more accurate and detailed 3D reconstructions.}
\label{fig:main-1}
\end{figure}

\begin{figure*}[!htb]
\begin{center}
\centerline{\includegraphics[width=1.0\textwidth]{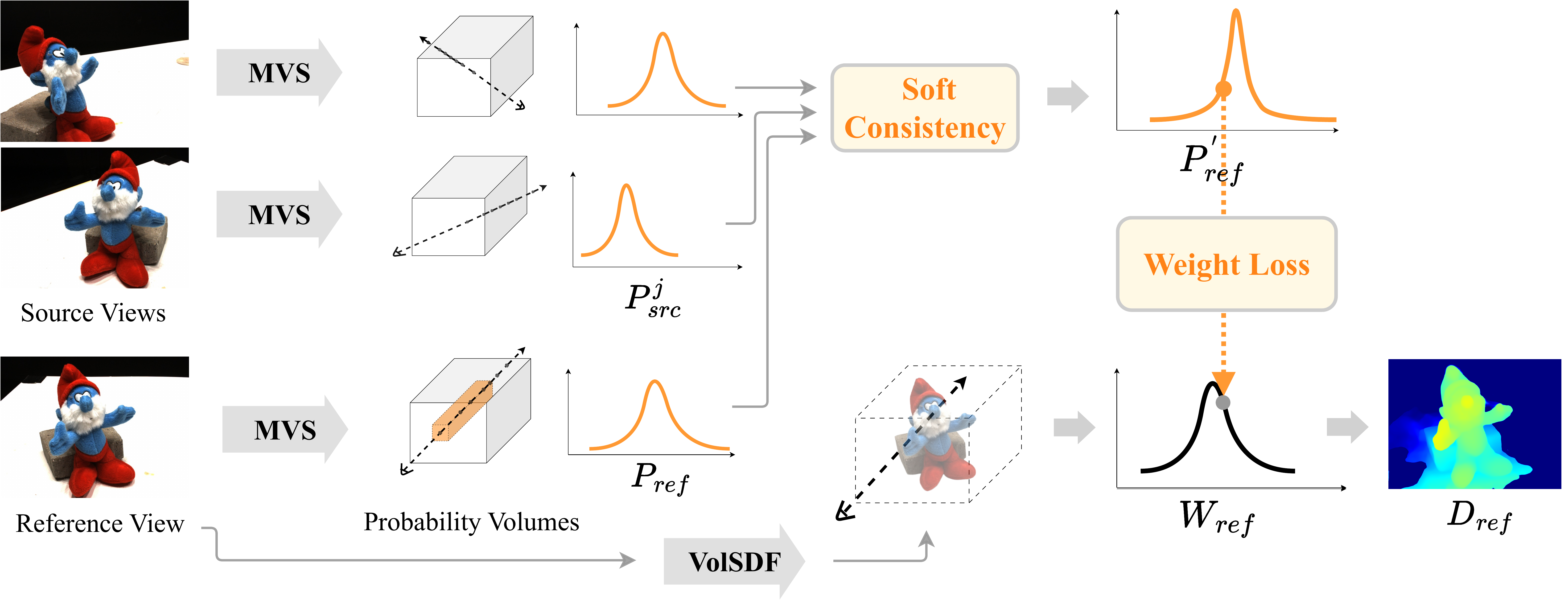}}
\end{center}
\caption{\textbf{Overview}. We propose to use probability volumes, obtained from multi-view stereo (MVS) models, to supervise the rendering weight estimated by VolSDF \cite{yariv2021volume}. We apply a soft consistency check to refine the volumes. The weight loss function ensures consistency between the probability volume and the rendering weight. This process allows us to use the reconstructed depth information provided by VolSDF to guide the depth hypothesis sampling in the MVS models, as depicted in \cref{fig:main-1}.}
\label{fig:main-2}
\end{figure*}

We propose a novel way to integrate neural volume rendering with multi-view stereo algorithms. Specifically, we adopt VolSDF \cite{yariv2021volume} for the neural surface reconstruction and notice that with sparse input views, VolSDF's reconstruction quality degrades dramatically. To mitigate this, we propose \MyMethod{} that makes use of the correspondence-aware probability volume from MVS algorithms. \cref{fig:main-1} and \cref{fig:main-2} provide an overall illustration of our method.

\subsection{Background}

\noindent \textbf{Volume Rendering of Implicit Surfaces.} We use forward volume rendering \cite{max1995optical, kajiya1984ray, mildenhall2020nerf} as our differentiable volumetric representation of the 3D scene and apply VolSDF \cite{yariv2021volume}. VolSDF represents scene geometry as a signed distance function (SDF), which is subsequently transformed into density values for volume rendering. For each pixel, we sample points between the near and far depths along the ray $\mathbf{r}$ and approximate the pixel color $\hat{C}$ by:
\begin{equation}
\begin{aligned}
\label{nerf_wi}
\hat{C}(\mathbf{r})&=\sum_{i=1}^{N} \myw_i \cdot \mathbf{c}_{i}, \\
\textrm{where} \quad \myw_i&=T_{i}\left(1-\exp \left(-\sigma_{i} \delta_{i}\right)\right), \\
T_{i}&=\exp \left(-\sum_{j=1}^{i-1} \sigma_{j} \delta_{j}\right).
\end{aligned}
\end{equation}
Here, $\myw_i$ is the rendering weight, $\sigma_{i}$ and ${c}_{i}$ denote the density and color at the sampled point $i$, respectively and $\delta_{i}$ is the distance between adjacent samples along the ray. The density value is approximated from the SDF $s$, with learnable parameter $\alpha, \beta$, as follows:
\begin{equation}
\begin{aligned}
\sigma(s)= \begin{cases}\frac{1}{2} \exp \left(\frac{s}{\beta}\right) \cdot \alpha & \text { if } s \leq 0 \\ \left(1-\frac{1}{2} \exp \left(-\frac{s}{\beta}\right)\right) \cdot \alpha & \text { if } s>0\end{cases}
\end{aligned}
\end{equation}

\subsection{\MyMethod}\label{ncf}

Implicit neural 3D representations usually require dense images, since their per-scene optimization can be seen as a trial-and-error process to determine the underlying 3D structures. Therefore, given sparse training views as supervision, neural rendering models often fit the training views flawlessly while the underlying geometry can be vastly incorrect \cite{niemeyer2022regnerf}. This can be understood as a local optimum and is known as the shape-radiance ambiguity problem \cite{zhang2020nerf++}. In Fig. \ref{fig:nerf_demo}, we demonstrate it experimentally by training VolSDF \cite{yariv2021volume} using 3 views only. As shown in Fig. \ref{fig:nerf_demo}, VolSDF completely fails to estimate geometry.

We propose to combine information from MVS to make neural rendering models correspondence-aware. The challenge lies in effectively incorporating noisy MVS predictions into VolSDF \cite{yariv2021volume}. We propose two steps:

\noindent \textbf{Soft Consistency.} Instead of the hard consistency constraints imposed by estimating the depth of each point, we impose soft consistency constraints by operating directly on the probability volumes: In MVS, depth maps are typically obtained by applying \emph{argmax} on the probability volume along each view direction. Then, photometric and geometric consistency checks \cite{yao2018mvsnet} are used to filter out depth outliers before fusing the depth maps into a point cloud. \emph{argmax} works well when dense inputs are available, but in the case of sparse inputs, the correct depth is often not assigned the highest probability. As a result, incorrect depths introduced by \emph{argmax} will be filtered out by consistency checks, resulting in an incomplete reconstruction. 

Alternatively, we propose directly computing consistency measures on the probability volumes. The \emph{reference} view is the image, the depth of which we want to determine. The other images are the \emph{source} views. By applying MVS to these views, we obtain probability volumes. Then, we multiply each probability value $\myP_{ref}(\myx)$ in the reference probability volume with 3D position $\myx$, with the sum of $\myP_{src}^j(\myx)$ at the same location, to compute a new consistency weighted probability volume. $\myP_{src}^j(\myx)$ is interpolated from the probability volumes of the source views. We demonstrate that this multiplication works adequately in our ablation study in \cref{sec: Ablation Study}. However, significant errors in depth still appear in challenging sparse-input scenarios.

\noindent \textbf{Noise-Tolerant Loss.} We further propose a noise-tolerant weight loss that utilizes the noisy probability volume to improve the reconstruction of VolSDF \cite{yariv2021volume}. Given points sampled along a viewing ray, we notice that $\myP$ and $\myw$ in \cref{nerf_wi} actually have the same meaning: their normalized values along the ray/depth both form a depth probability mass function, which can also be seen as \emph{the probability that a correspondence exists}. The larger the value of $\myw$, the more likely it is that this point is visible in multiple views (i.e. a correspondence between pixels in different images). Instead of directly checking for consistency between the different probability volumes, we use them (in the form of $\myw$) during neural rendering optimization. Thus, we leverage the smoothness of neural implicit models and combine the global consistency guaranteed by volumetric rendering. 

Specifically, we use our consistency weighted probability volume as supervision to regularize $\myw$ in the volume rendering \cref{nerf_wi}. Based on \cite{peng2022rethinking}, we can think of all possible 3D points on a ray as interior or exterior to the object (i.e. a binary classification problem). Thus, the MVS probability volume becomes a set of noisy positive labels for the rendering weights (i.e. occupancy values) with confidence from soft consistency. Hence, we have a classification problem that allows the use of cross entropy loss to optimize neural rendering methods. However, as shown in \cite{zhang2018generalized, song2022learning}, the cross entropy loss is sensitive to noisy labels. Based on insights from \cite{zhang2018generalized}, we adopt a generalized cross entropy loss in \cref{weight_loss} to reduce the penalty on false positive MVS predictions and thus increase optimization tolerance to noise. The noise tolerance level can be controlled by parameter $q$, where the generalized cross entropy loss is equivalent to the cross entropy loss when $q$ approaches $0$ \cite{zhang2018generalized}, and to the Mean Absolute Error (MAE) loss when $q=1$. Our noise-tolerant weight loss is shown in \cref{weight_loss}.
\begin{equation}
\begin{aligned}
\mathcal{L}_{weight} &= \sum_{\myx \in \mathcal{X}} \frac{1-{\myw(\myx)}^q}{q} \cdot \myP_{ref}^{\prime}(\myx), \\
\textrm{where} \quad
\myP_{ref}^{\prime}(\myx) &= \mathbf{P}_{ref}(\myx) \cdot \sum_{j} \myP_{src}^j(\myx),
\label{weight_loss}
\end{aligned}
\end{equation}
\noindent $\myw(\myx)$ is the rendering weight predicted by the neural rendering model at the sampled location $\myx$ along a ray in the reference view, $\myP_{ref}$ is the probability volume in the reference view, and $\myP_{src}^j$ is the probability volume of a source view. In this way, we are essentially optimizing correspondences across images in a globally consistent and noise-tolerant way. In spirit, this is similar to finding a graph-cut in a volume of correspondence costs described in \cite{vogiatzis2005multi}.

\noindent \textbf{Coarse-to-fine MVS Reconstruction.} As shown in \cref{fig:main-1}, we incorporate our method into three coarse to fine MVS models \cite{ding2022transmvsnet, cheng2020deep, gu2020cascade}. We use the first coarse stage MVS probability volume to guide VolSDF \cite{yariv2021volume} optimization. After that, we use the depth map obtained from VolSDF and replace the original depth map estimated by the coarse stage MVS model to remove the noise in the coarse stage MVS depth. We then follow the same protocol as in coarse-to-fine MVS models: use the depth map estimated from the coarse stage to guide the sampling range of the depth candidates of the next, finer stage in MVS. Because our coarse guidance depth map is more complete and accurate, the next stage MVS depth estimation is simpler. Therefore, we only use half of the depth search width in the finer stages compared to the default search width used in MVS models. Our surface reconstruction is more complete and still accurate compared to MVS models. Furthermore, as we show in \cref{tab: dtu_chamfer}, our method can be effortlessly incorporated into most coarse-to-fine MVS models and achieves considerable improvement compared to standalone MVS models.

\noindent \textbf{Optimization.} We use the same loss functions as VolSDF \cite{yariv2021volume}, along with our weight loss and sparsity regularization:
\begin{equation}
\begin{aligned}
\mathcal{L}_{\text {sparse }}=\frac{1}{\|\mathbb{Q}\|} \sum_{r \in \mathbb{Q}} \ 1 / (d_r + \epsilon),
\label{eq: sparse_loss}
\end{aligned}
\end{equation}
\noindent where $d_r$ are predicted depths and $\mathbb{Q}$ are rays without MVS supervision ($\sum \myPprime(\myx) \approx 0$). We encourage sparsity by maximizing depth values. $\mathcal{L}_{\text {sparse}}$ is only used in the first 200 steps, along with heavily Gaussian-smoothed images as photometric supervision to suppress floating surfaces.

\noindent \textbf{Rendering.} For coordinate-based MLPs, fitting high-frequency details and maintaining good geometry simultaneously is challenging \cite{yariv2021volume}. Since our method produces reasonably good geometry, we experiment with a simple image-based rendering approach \cite{debevec1996modeling, chen1993view, buehler2001unstructured} in testing to warp nearby view pixels based on predicted depth maps to synthesize novel views. In areas where there are no valid pixels to warp (i.e., the geometric consistency check between the rendered depths of the novel view and input views fails), we use rendered colors. A 4-level Laplacian pyramid \cite{burt1987laplacian} is used to smoothly blend the warped pixels. Our method with image-based rendering is denoted as \textbf{Ours$_{IR}$}.

\section{Experiments}
\label{sec: Experiments}

We evaluate our method on complex multi-view 3D surface reconstruction tasks, using two datasets: DTU \cite{aanaes2016large} and BlendedMVS \cite{yao2020blendedmvs}, both featuring real objects with diverse materials captured from multiple views. We demonstrate the superiority of our approach over prior work through quantitative and qualitative evaluation (\cref{sec: Comparisons}). Furthermore, we conduct extensive ablation studies to verify the effectiveness of our design choices (\cref{sec: Ablation Study}).

\subsection{Experimental Settings}
\label{sec: Experimental Settings}

\noindent \textbf{Datasets.}
For the DTU dataset \cite{aanaes2016large}, we combine the scans used in \cite{yariv2020multiview, yariv2021volume, yu2021pixelnerf} with the ones used in conventional MVS settings \cite{ding2022transmvsnet, yao2018mvsnet}, and remove the training scans of common MVS models. Our primary experiments are on three-view 3D reconstruction. Similar to PixelNeRF \cite{yu2021pixelnerf}, we use views 25, 22, and 28 for three-view reconstruction. We further test on 6 and 9 input views with consistent improvements in performance\footnote{Results for the 6 and 9 image scenarios are in supplementary.}. For the BlendedMVS dataset \cite{yao2020blendedmvs}, we select 9 challenge scenes, following \cite{yariv2021volume}. For each scene, we select a set of sparse input views (i.e. 3 images) with a relatively wide baseline, similar to the setting in the DTU dataset. The image resolution is set to 768 $\times$ 576 for both the DTU and BlendedMVS datasets. We use foreground masks from \cite{niemeyer2022regnerf, yariv2020multiview} following \cite{niemeyer2022regnerf, long2022sparseneus} for evaluation.


\noindent \textbf{Metrics.}
For surface reconstruction, we follow the standard evaluation protocol in \cite{aanaes2016large, yariv2020multiview, yariv2021volume} and report the Chamfer distance (in mm) of the output point clouds with ground truth point clouds. For novel view  synthesis, we adopt the mean of PSNR, structural similarity index (SSIM) \cite{wang2004image}, and the LPIPS perceptual metric \cite{zhang2018unreasonable}. 

\noindent \textbf{Implementation details.} We experiment mainly using CasMVSNet \cite{gu2020cascade} to obtain the cascade probability volume. We notice that, given only 3 input views, the default plane sweep settings (48, 32, and 8 depth hypotheses with interval widths 4, 2, and 1 respectively) do not retain fine details very well. We change them to 192, 32, and 8 depth hypotheses with intervals 1, 0.5, and 0.5 respectively. We are able to make the finer stage depth search interval widths much smaller because our method produces more complete and accurate coarse depth maps. The batch size is 512 rays. The $q$ in our weight loss \cref{weight_loss} is 0.5 in all experiments. We optimize each scene for 100K steps. Before fusing the depth maps output by the MVS model into a point cloud, standard photometric and geometric consistency \cite{yao2018mvsnet} checks based on probability values and depth errors are adopted.


\begin{table*}[!htb]\centering
\centering
\begin{tabular}{lcccccccccccc}
\toprule
Scan  & 21   & 24  & 34   & 37  & 38   & 40   & 82   & 106  & 110  & 114  & 118  & \textbf{Mean} \\
\midrule
IBRNet$_{ft}$ \cite{wang2021ibrnet} & 3.40 & 3.54 & 3.13 & 6.78 & 3.32 & 4.80 & 3.48 & 2.59 & 3.93 & 1.23 & 2.74 & 3.54 \\
MVSNeRF \cite{chen2021mvsnerf} & 2.07 & 2.35 & 1.23 & 3.87 & 1.36 & 2.40 & 2.23 & 1.64 & 1.76 & 0.65 & 1.86 & 1.95 \\
GeoNeRF \cite{johari2022geonerf} & 2.13 & 3.04 & 1.00 & 3.93 & 1.20 & 2.46 & 2.32 & 1.82 & 2.21 & 0.79 & 1.67 & 2.06 \\
SparseNeuS$_{ft}$ \cite{long2022sparseneus} & 5.26 & 4.93 & 5.59 & 7.04 & 5.18 & 7.38 & 4.78 & 4.58 & 5.61 & 4.55 & 5.61 & 5.51 \\
NeuS \cite{wang2021neus} & 4.52 & 3.33 & 3.03 & 4.77 & 1.87 & 4.35 & 1.89 & 4.18 & 5.46 & 1.09 & 2.40 & 3.36 \\
VolSDF \cite{yariv2021volume} & 4.54 & 2.61 & 1.51 & 4.05 & 1.27 & 3.58 & 3.48 & 2.62 & 2.79 & 0.52 & 1.10 & 2.56 \\
\midrule
TransMVSNet \cite{ding2022transmvsnet} & 3.39 & 3.61& 1.55 & 4.24& 1.95 & 3.06 & 3.45 & 2.94 & 3.81 & 1.67 & 2.34 & 2.92 \\
\cite{ding2022transmvsnet} + \textbf{Ours}  & 1.91 & 2.08& 0.98 & 2.94& 1.21 & 1.90  & 2.70  & 1.56 & 1.99 & 1.09 & 1.35 & 1.80 \\
\cmidrule(l{0.7em}r{0.7em}){1-13}
UCSNet \cite{cheng2020deep}       & 2.57 & 3.01& 1.82 & 4.07& 1.62 & 3.10 & 2.49 & 1.93 & 1.27 & 0.68 & 1.59 & 2.20 \\
\cite{cheng2020deep} + \textbf{Ours}       & \textbf{1.89} & 2.12& 1.24 & 3.17& 1.07 & 2.07 & 1.38 & \textbf{1.24} & 0.78 & 0.54 & 1.16 & 1.52 \\
\cmidrule(l{0.7em}r{0.7em}){1-13}
CasMVSNet{\cite{gu2020cascade}}   & 2.40  & 3.07& 1.23 & 3.27& 1.35 & 2.76 & 1.82 & 1.72 & 1.30  & 0.70  & 1.44 & 1.92 \\
\cite{gu2020cascade} + \textbf{Ours}   & 1.96 & \textbf{1.99} & \textbf{0.74} & \textbf{2.58} & \textbf{0.95} & \textbf{1.47} & \textbf{1.37} & 1.32 & \textbf{0.54} & \textbf{0.51} & \textbf{1.03} & \textbf{1.32} \\
\bottomrule
\end{tabular}
\caption{Quantitative results on 3D reconstruction for the DTU dataset. "+ Ours" means that we use the cited MVS algorithm as the probability volume builder and optimize using our method. The metric is the Chamfer distance (lower is better).}
\label{tab: dtu_chamfer}
\end{table*}

\begin{figure*}[!htb]
\begin{center}
\includegraphics[width=1.0\textwidth]{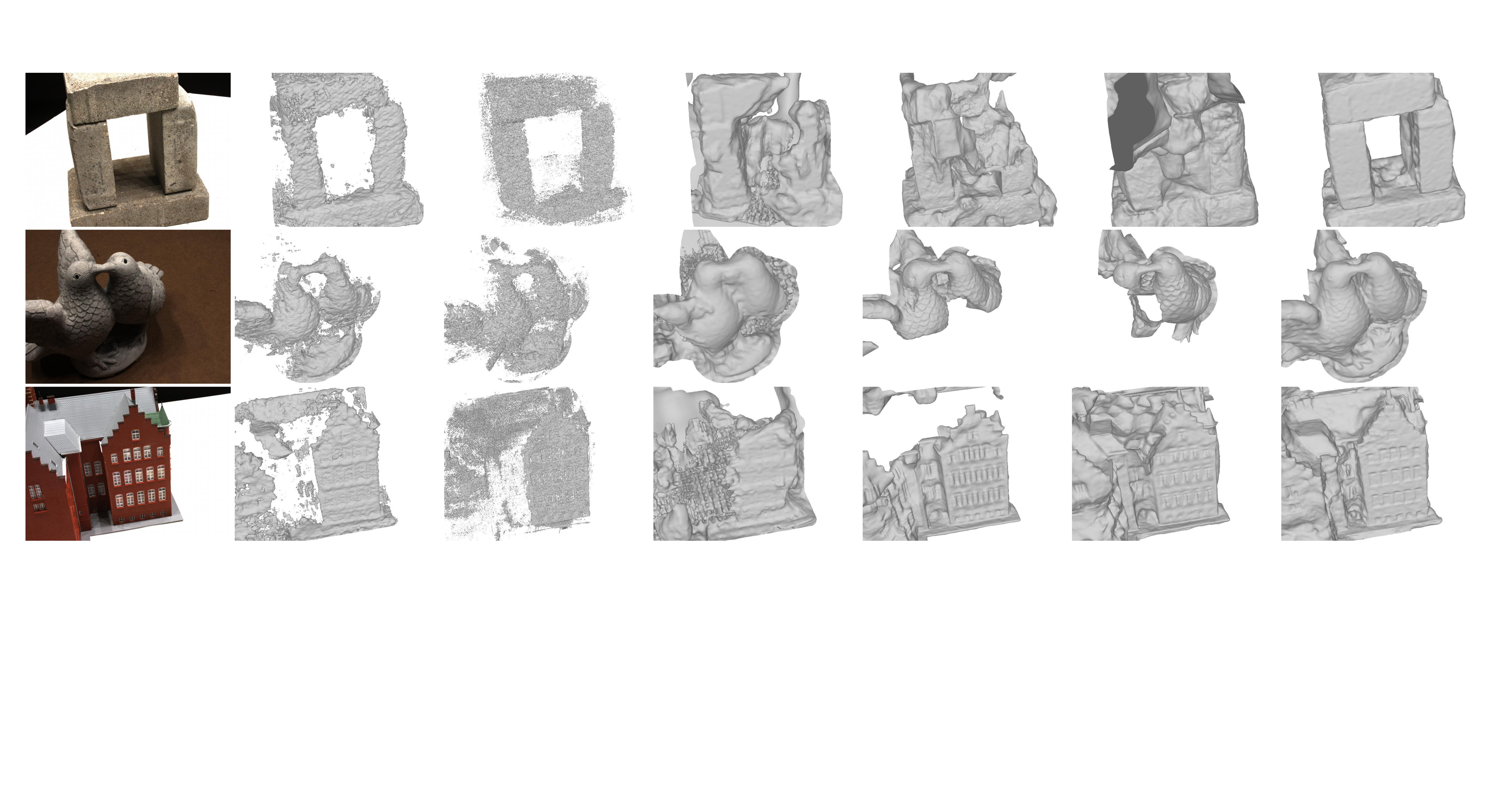}
\leftline{\hspace{0.05\textwidth} GT \hspace{0.06\textwidth} MVSNeRF \cite{chen2021mvsnerf} \hspace{0.015\textwidth} GeoNeRF \cite{johari2022geonerf} \hspace{0.005\textwidth} SparseNeuS$_{ft}$ \cite{long2022sparseneus} \hspace{0.01\textwidth} NeuS \cite{wang2021neus} \hspace{0.04\textwidth} VolSDF \cite{yariv2021volume} \hspace{0.05\textwidth} \textbf{Ours}}
\end{center}
\caption{3D reconstruction results of neural rendering methods on DTU. Our results appear more complete and accurate. 
}
\label{fig:dtu_mesh}
\end{figure*}

\subsection{Baselines}
\label{sec: Baselines}

\noindent \textbf{Neural Rendering Methods.} We compare against state-of-the-art generic neural rendering methods, including IBRNet \cite{wang2021ibrnet}, MVSNeRF \cite{chen2021mvsnerf}, GeoNeRF \cite{johari2022geonerf}, and SparseNeuS \cite{long2022sparseneus}. We fine-tune IBRNet and SparseNeuS using three input images for each scene for 20K and 10k iterations, respectively. We only report non-fine-tuned results for MVSNeRF and GeoNeRF because our attempts to fine-tune using 3 images did not succeed due to the inherent difficulty of the task, consistent with \cite{niemeyer2022regnerf, long2022sparseneus}. Additionally, we compare our method with per-scene optimization based neural surface reconstruction methods, NeuS \cite{wang2021neus} and VolSDF \cite{yariv2021volume}, with VolSDF being the neural rendering model used to refine MVS predictions in our method. For fair comparison with MVS, we only maintain the foreground depth maps generated by neural rendering techniques by applying standard geometric consistency checks and ground truth masks. We merge the depth maps into a point cloud for evaluation \cite{yao2018mvsnet}.

\noindent \textbf{MVS Methods.} To evaluate the generalizability of our method, we incorporate it into three state-of-the-art coarse to fine MVS models: CasMVSNet \cite{gu2020cascade}, UCSNet \cite{cheng2020deep}, and TransMVSNet \cite{ding2022transmvsnet}. All MVS networks are pre-trained only on DTU \cite{aanaes2016large} with ground-truth depth as supervision and are frozen during per-scene optimization.

\subsection{Comparisons}
\label{sec: Comparisons}


\begin{figure*}[!htb]
\begin{center}
\includegraphics[width=1.0\textwidth]{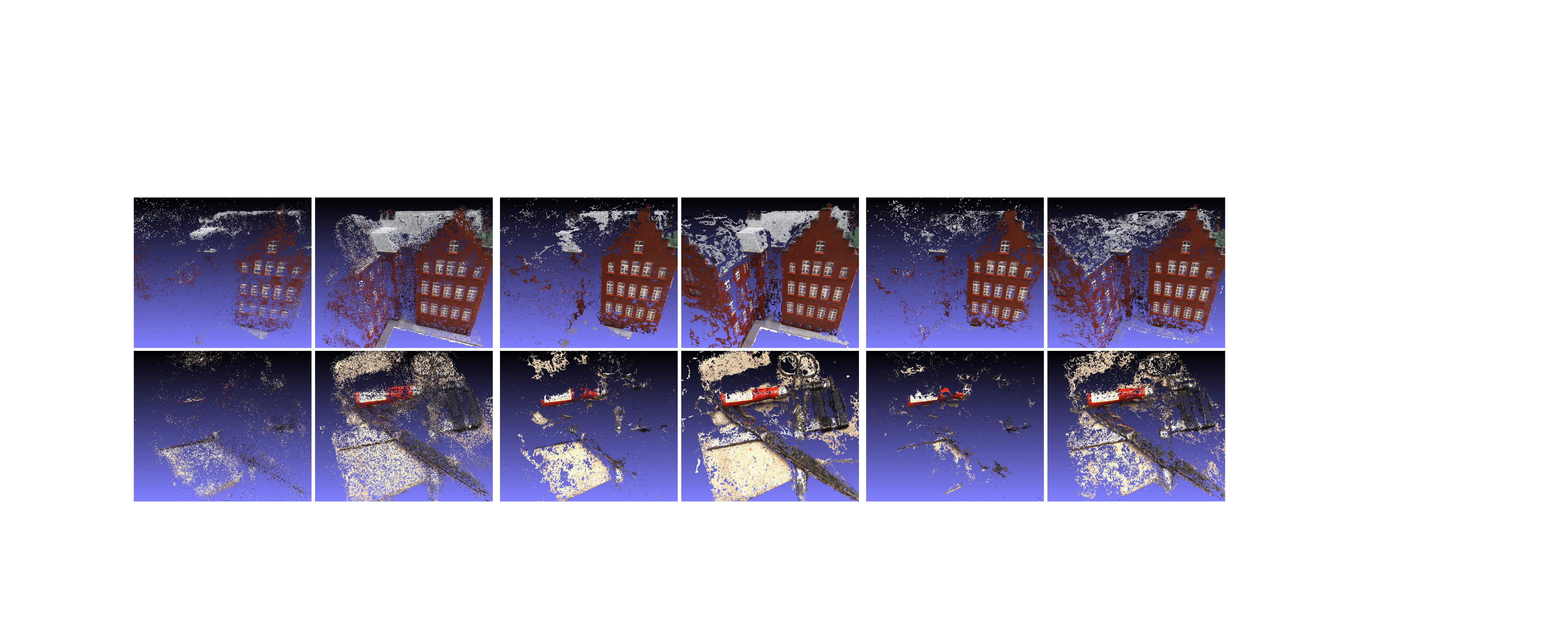}
\leftline{\hspace{0.01\textwidth} TransMVSNet \cite{ding2022transmvsnet} \hspace{0.02\textwidth} \cite{ding2022transmvsnet}+Ours \hspace{0.055\textwidth}  CasMVSNet \cite{gu2020cascade} \hspace{0.05\textwidth}  \cite{gu2020cascade}+Ours \hspace{0.06\textwidth}  UCSNet \cite{cheng2020deep} \hspace{0.06\textwidth}  \cite{cheng2020deep}+Ours}
\end{center}
\caption{Point cloud visualization on DTU. Results improve in all combinations of our method with different MVS models.}
\label{fig: dtu_pcd2}
\end{figure*}

\begin{figure*}[!htb]
\begin{center}
\includegraphics[width=1.0\textwidth]{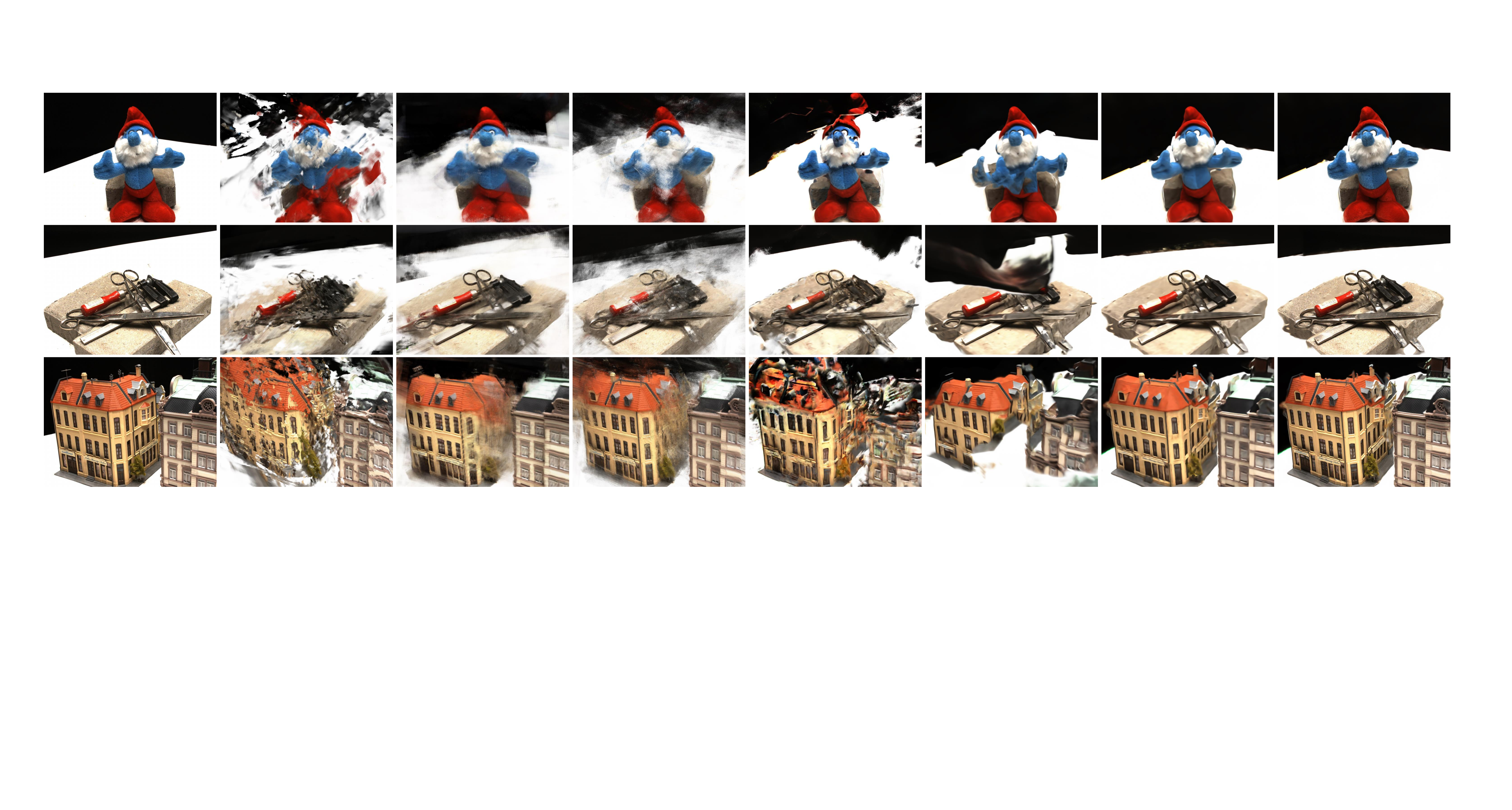}
\leftline{
\hspace{0.04\textwidth}    GT    \hspace{0.05\textwidth}  IBRNet$_{ft}$ \cite{wang2021ibrnet} \hspace{0.0\textwidth} MVSNeRF \cite{chen2021mvsnerf} \hspace{0.0\textwidth} GeoNeRF \cite{johari2022geonerf} \hspace{0.00\textwidth}  NeuS \cite{wang2021neus}    \hspace{0.02\textwidth} VolSDF \cite{yariv2021volume}  \hspace{0.04\textwidth}    \textbf{Ours}            \hspace{0.065\textwidth}     \textbf{Ours$_{IR}$}}
\end{center}
\caption{Our method appears to be more accurate in novel view synthesis on DTU.}
\label{fig:dtu_nvs}
\end{figure*}

\begin{table*}[!htb]\centering
\begin{tabular}{lcccccccccc}
\toprule
Scene & Doll & Egg & Head & Angel & Bull & Robot & Dog & Bread & Camera & \textbf{Mean} \\
\midrule
MVSNeRF \cite{chen2021mvsnerf}   & 5.3  & -16.8 & -17.7 & 38.2 & 13.8 & 11.9 & -0.3 & 12.3 & 8.0  & 6.1 \\
GeoNeRF \cite{johari2022geonerf} & 29.3 & 21.4 & 11.5 & 37.6 & -0.6 & -15.1 & 13.7 & 21.4 & 11.5 & 14.5 \\
\cmidrule(l{0.7em}r{0.7em}){1-11}
CasMVSNet  \cite{gu2020cascade}  & 32.9 & 47.1 & 17.3 & 45.9 & 11.3 & 11.5 & 33.3 & 19.2 & 30.1 & 27.6 \\
\textbf{Ours} & \textbf{35.0} & \textbf{58.8} & \textbf{38.5} & \textbf{54.7} & \textbf{33.4} & \textbf{23.9} & \textbf{33.7} & \textbf{64.4} & \textbf{43.4} & \textbf{42.9} \\
 \bottomrule
\end{tabular}
\caption{BlendedMVS 3D reconstruction results. Since there are no units in BlendedMVS, we report relative improvement (in \%) over VolSDF \cite{yariv2021volume} in terms of Chamfer distance.}
\label{bmvs_chamfer}
\end{table*}

\begin{figure*}[htb]
\begin{center}
\includegraphics[width=1.0\textwidth]{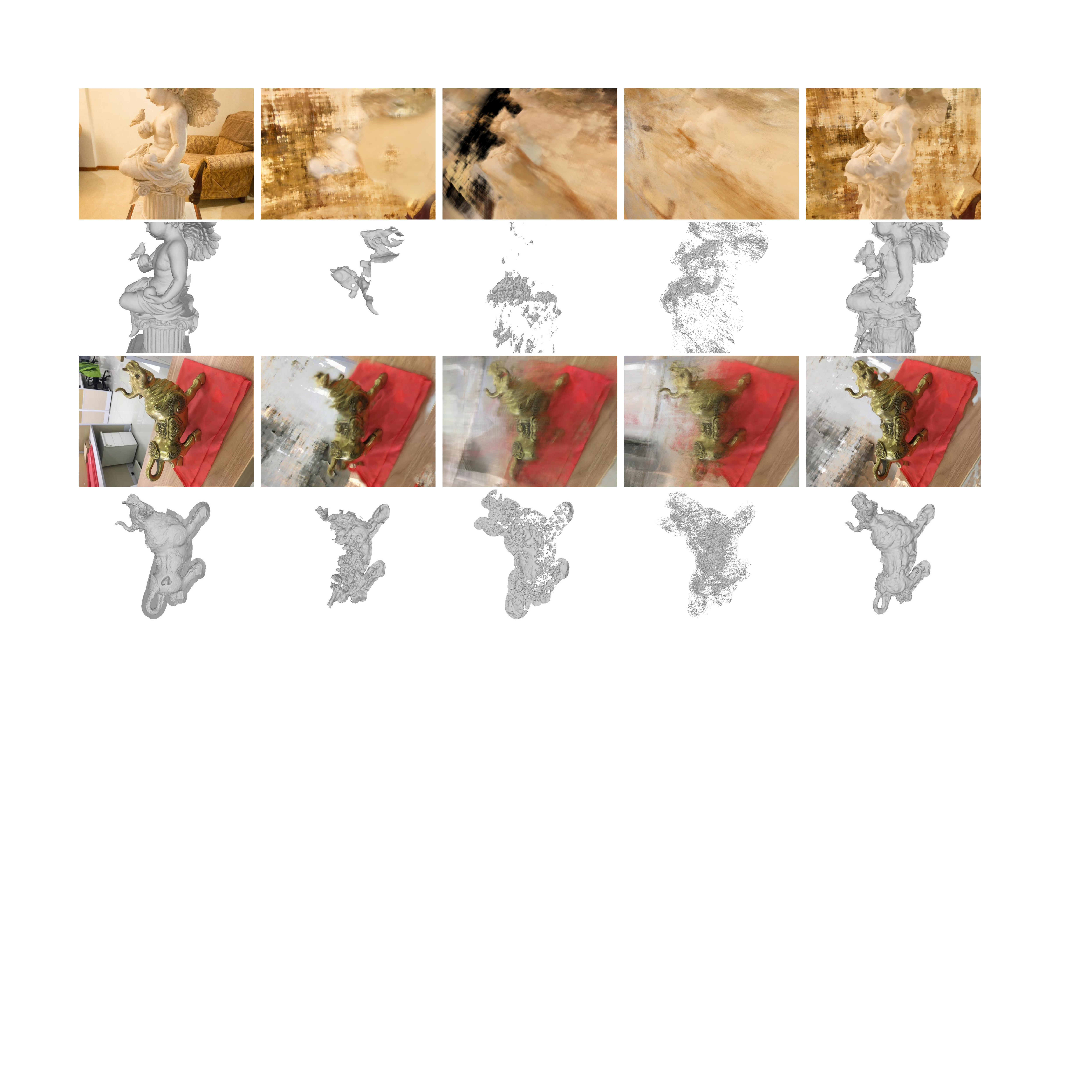}
\leftline{
\hspace{0.07\textwidth}   GT
\hspace{0.14\textwidth} VolSDF \cite{yariv2021volume} \hspace{0.07\textwidth} MVSNeRF \cite{chen2021mvsnerf} \hspace{0.07\textwidth} GeoNeRF \cite{johari2022geonerf} \hspace{0.1\textwidth} \textbf{Ours$_{IR}$}}
\end{center}
\caption{3D reconstruction and novel view synthesis comparisons on BlendedMVS. Our results appear more complete and accurate.}
\label{fig: bmvs_mesh}
\end{figure*}

\noindent \textbf{3D Reconstruction.} Our approach surpasses state-of-the-art techniques in 3D reconstruction, as demonstrated by its superior performance on both the DTU \cite{aanaes2016large} and BlendedMVS datasets \cite{yao2020blendedmvs} (\cref{tab: dtu_chamfer} and \cref{bmvs_chamfer}). We show  meshes extracted from neural rendering method outputs in \cref{fig:dtu_mesh} and \cref{fig: bmvs_mesh}.

\begin{figure}[htb]
\begin{center}
\includegraphics[width=0.45\textwidth]{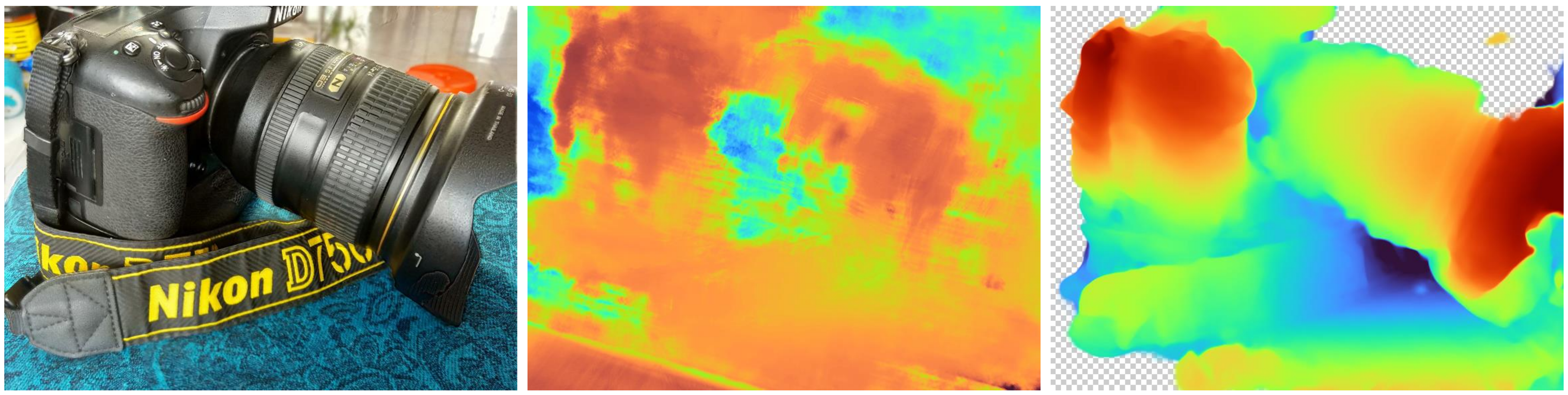}
\leftline{\hspace{0.085\textwidth} GT \hspace{0.05\textwidth}  MVSNeRF \cite{chen2021mvsnerf} \hspace{0.05\textwidth}  Ours}
\end{center}
\caption{Depth map predictions on BlendedMVS using MVSNeRF \cite{chen2021mvsnerf} and our method. Improved depths are an illustration of better geometry-appearance disentanglement.}
\label{fig: bmvs_normal}
\end{figure}

As shown in \cref{fig:dtu_mesh}, VolSDF \cite{yariv2021volume} and NeuS \cite{wang2021neus} show suboptimal performance due to the weak photometric constraint in resolving the shape-radiance ambiguity. Fine-tuning SparseNeuS \cite{long2022sparseneus} can lead to degenerate results, especially on the BlendedMVS dataset, so we only report its performance on DTU. Fine-tuned IBRNet \cite{wang2021ibrnet} performs worse than methods using stronger MVS priors such as MVSNeRF \cite{chen2021mvsnerf} and GeoNeRF \cite{johari2022geonerf}. Although MVSNeRF and GeoNeRF demonstrate impressive performance, they still fall short compared to our method (see \cref{fig: bmvs_normal}). 



As shown in \cref{tab: dtu_chamfer} and \cref{fig: dtu_pcd2}, MVS models coupled with our noise-tolerant optimization perform much better than MVS models or VolSDF \cite{yariv2021volume} alone. Thus, our method can be treated as a general module that can be plugged into other MVS methods and improve their performance.

With the introduction of MVS information, we enable fast per-scene surface optimization. Our output surface reconstruction after 10-15 minutes of training (on an NVIDIA A5000 GPU) is already better than the reconstruction of the fully trained VolSDF \cite{yariv2021volume} (typically 4-10 hours). More specifically, on DTU, we obtain 39\% better Chamfer distance over the fully-trained VolSDF after 15 minutes of optimization, with our final model achieving a 48\% improvement. Please refer to the supplementary for more details.

\noindent \textbf{Novel View Synthesis.} Our method excels at improving geometry, yet also demonstrates competitive performance in novel view synthesis (as shown in \cref{tab: dtu_nvs} and \cref{tab: bmvs_nvs}). \cref{fig:dtu_nvs} and \cref{fig: bmvs_mesh} illustrate improved view synthesis results compared to other methods, suggesting our method's capacity to better disentangle geometry and texture. Also, adding the image interpolation to the rendering process greatly enhances LPIPS, while slightly improving PSNR and SSIM, by incorporating more details, as demonstrated in \cref{tab: dtu_nvs}, \cref{tab: bmvs_nvs}, and \cref{fig:dtu_nvs}.

\begin{table}[htb]\centering
\begin{tabular}{lccc}
\toprule
\textbf{Method} & PSNR $\uparrow$ & SSIM $\uparrow$ & LPIPS $\downarrow$ \\
\midrule
IBRNet$_{ft}$ \cite{wang2021ibrnet} & 15.71 & 0.759 & 0.295 \\
MVSNeRF \cite{chen2021mvsnerf} & 18.37 & 0.818 & 0.254 \\
GeoNeRF \cite{johari2022geonerf} & 19.45 & 0.837 & 0.220 \\
NeuS \cite{wang2021neus} & 15.34 & 0.753 & 0.313 \\
VolSDF \cite{yariv2021volume} & 16.99 & 0.786 & 0.332 \\
\textbf{Ours} & 20.21 & 0.820 & 0.321 \\
\textbf{Ours$_{IR}$} & \textbf{20.58} & \textbf{0.855} & \textbf{0.157} \\
\bottomrule
\end{tabular}
\caption{Novel view synthesis comparisons on DTU.}
\label{tab: dtu_nvs}
\end{table}

\begin{table}[htb]\centering
\begin{tabular}{lccc}
\toprule
\textbf{Method} & PSNR $\uparrow$ & SSIM $\uparrow$ & LPIPS $\downarrow$ \\
\midrule
MVSNeRF \cite{chen2021mvsnerf} & 14.99 & 0.866 & 0.164 \\
GeoNeRF \cite{johari2022geonerf} & 17.09 & 0.886 & 0.139 \\
VolSDF \cite{yariv2021volume} & 14.47 & 0.860 & 0.182 \\
\textbf{Ours} & 16.97 & 0.893 & 0.154 \\
\textbf{Ours$_{IR}$} & \textbf{17.26} & \textbf{0.906} & \textbf{0.105} \\
\bottomrule
\end{tabular}
\caption{Novel view synthesis comparisons for BlendedMVS.}
\label{tab: bmvs_nvs}
\end{table}

\subsection{Ablation Study}
\label{sec: Ablation Study}

We conduct ablation studies on the DTU dataset (\cref{ablate}). First, we show that using only the soft consistency constraints without additional optimization still improves the reconstruction result. This supports our assumption that the probability volumes contain more information than lossy depth maps obtained from an \emph{argmax} operation. Second, to evaluate the effectiveness of our weight loss, we replace our loss with the mean squared error (MSE) between the reconstructed depth from VolSDF and the geo-consistency filtered depth map obtained from MVS, similar to DS-NeRF \cite{deng2022depth}. Third, replacing the probability volumes with the depth maps as input, led to worse performance\footnote{We set the probability to be 1 only at the depth prediction location.}. Finally, we replace our weight loss with cross entropy loss, showing that generalized cross entropy loss is indeed noise-tolerant. Due to the trade-off between accuracy and completeness in point cloud filtering, we use Chamfer distance as the metric, following \cite{yariv2020multiview, yariv2021volume}. See supplementary for more details.

\begin{table}[htb]\centering
\begin{tabular}{lc}
\toprule
\textbf{Method} & Chamfer $\downarrow$ \\
\midrule
VolSDF \cite{yariv2021volume}  & 2.558 \\
CasMVSNet \cite{gu2020cascade}  & 1.920 \\
\textbf{Ours}           & \textbf{1.320} \\
\hspace{0.01\textwidth} only soft consistency         & 1.711 \\
\hspace{0.01\textwidth} MSE loss \cite{deng2022depth}  & 1.792 \\
\hspace{0.01\textwidth} w/o probability volume     & 1.543 \\
\hspace{0.01\textwidth} w/o GCE loss               & 1.534 \\
\bottomrule
\end{tabular}
\caption{Ablation studies for the DTU dataset. All rows except the first three are our model with different ablated components.}
\label{ablate}
\end{table}

\section{Conclusions}
\label{sec: Conclusions}

We presented \MyMethod{}, a novel approach to recover underlying geometry from sparse input views. Neural rendering optimization mainly relies on dense input images so that it can use trial-and-error mechanisms for reconstruction. Hence, its performance drops considerably with sparse inputs. We regularized the weight distribution with a refined probability volume obtained from MVS algorithms. We further made our method noise-tolerant by applying a generalized cross entropy loss. Our experiments show that our model not only outperforms neural rendering models but also significantly boosts the performance of MVS algorithms.

\noindent \textbf{Discussion and Limitations.}  While our method is capable of refining the probability volumes of the finer stages of MVS, we notice diminishing improvement because there is not much noise left in these stages. We include an ablation study on this in the supplementary material. While neural rendering models are able to deal with non-opaque, texture-less, or glossy surfaces, our introduction of MVS reduces this ability. This is an interesting area of research, particularly in the context of few-view reconstruction.

\vspace{\baselineskip}
\noindent \textbf{Acknowledgements.} This work was supported in part by the NASA Biodiversity Program (Award 80NSSC21K1027), and NSF Grant IIS-2212046. We also thank Alfredo Rivero for his thoughtful feedback and meticulous proofreading. 

\clearpage

{\small
\bibliographystyle{ieee_fullname}
\bibliography{egbib}
}


\appendix

\onecolumn

\setlength{\parskip}{1mm}

\begin{center}
    \Large \textbf{Appendix}
\end{center}

In \cref{sec: add} we report additional results on 3D reconstructions, novel view synthesis, the implicit surface optimization process, scalability, and limitations of our method. In \cref{sec: add_impl} we describe in further detail our experiment settings. We also include a supplementary video that compares the results of our method against various baselines.

\section{Additional Results}
\label{sec: add}

\begin{figure}[htb]
\begin{center}
\includegraphics[width=1.0\textwidth]{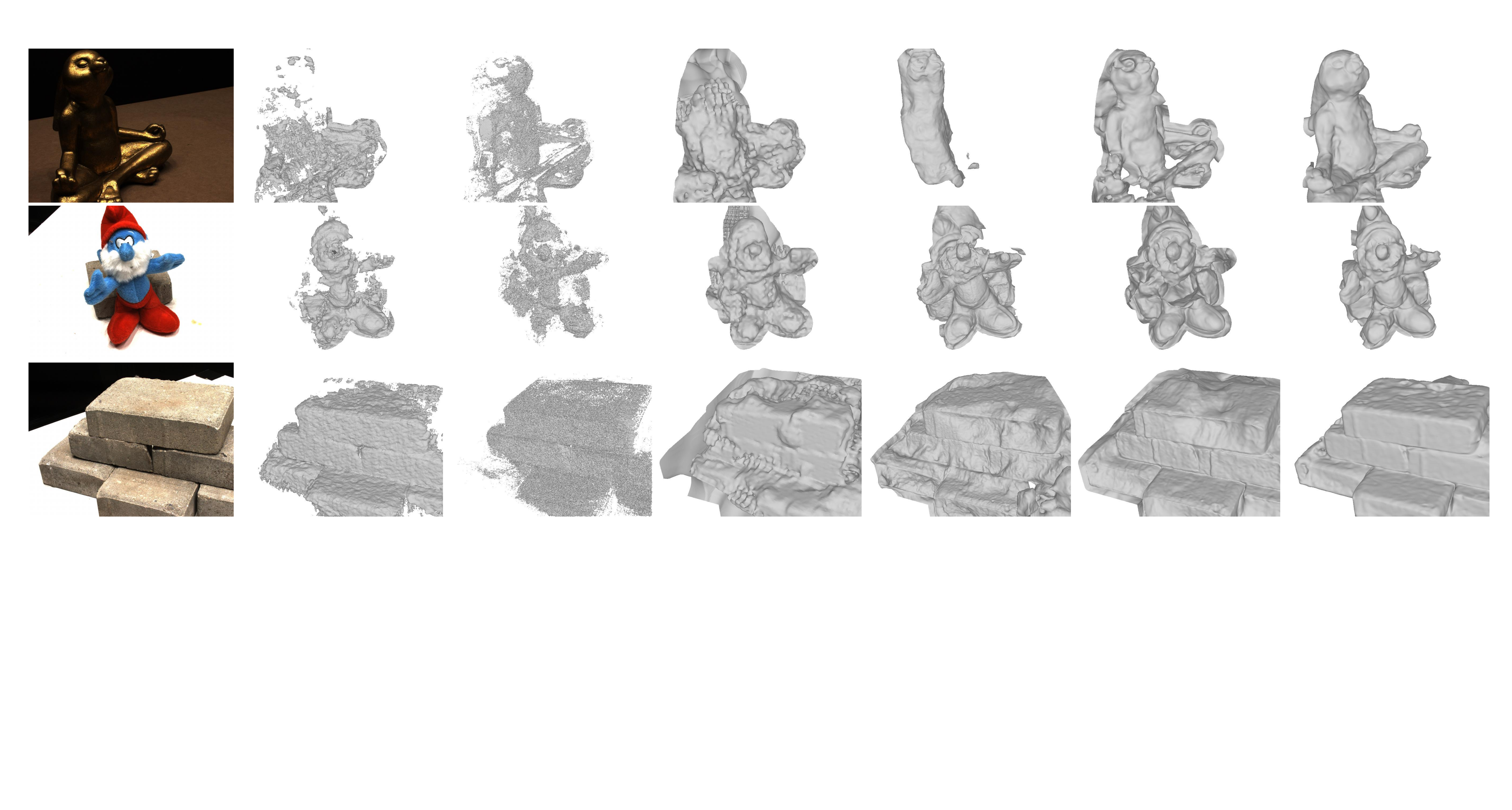}
\leftline{\hspace{0.06\textwidth} GT \hspace{0.06\textwidth} MVSNeRF \cite{chen2021mvsnerf} \hspace{0.015\textwidth} GeoNeRF \cite{johari2022geonerf} \hspace{0.005\textwidth} SparseNeuS$_{ft}$ \cite{long2022sparseneus} \hspace{0.015\textwidth} NeuS \cite{wang2021neus} \hspace{0.03\textwidth} VolSDF \cite{yariv2021volume} \hspace{0.045\textwidth} \textbf{Ours}}
\end{center}
\caption{Additional 3D reconstruction results of neural rendering methods on DTU. Our results appear more complete and accurate.
}
\label{fig: dtu_mesh_sup}
\end{figure}

\begin{figure}[htb]
\begin{center}
\includegraphics[width=1.0\textwidth]{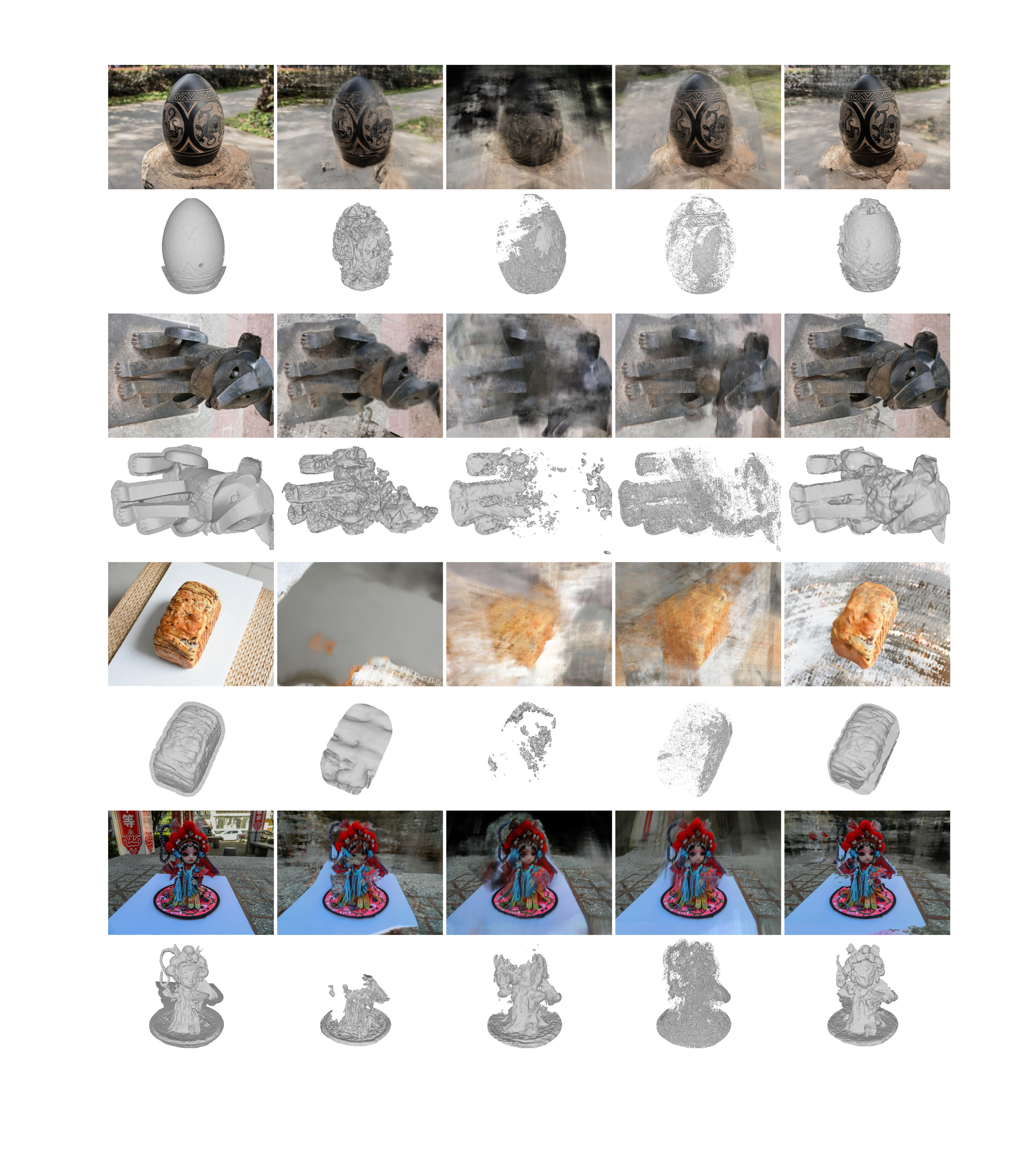}
\leftline{
\hspace{0.075\textwidth}   GT \hspace{0.125\textwidth} VolSDF \cite{yariv2021volume} \hspace{0.08\textwidth} MVSNeRF \cite{chen2021mvsnerf} \hspace{0.08\textwidth} GeoNeRF \cite{johari2022geonerf} \hspace{0.1\textwidth} \textbf{Ours$_{IR}$}}
\end{center}
\caption{Additional 3D reconstruction and novel view synthesis comparisons on BlendedMVS. Our results appear more complete and accurate.}
\label{fig: bmvs_mesh_sup}
\end{figure}

\begin{figure}[htb]
\begin{center}
\includegraphics[width=1.0\textwidth]{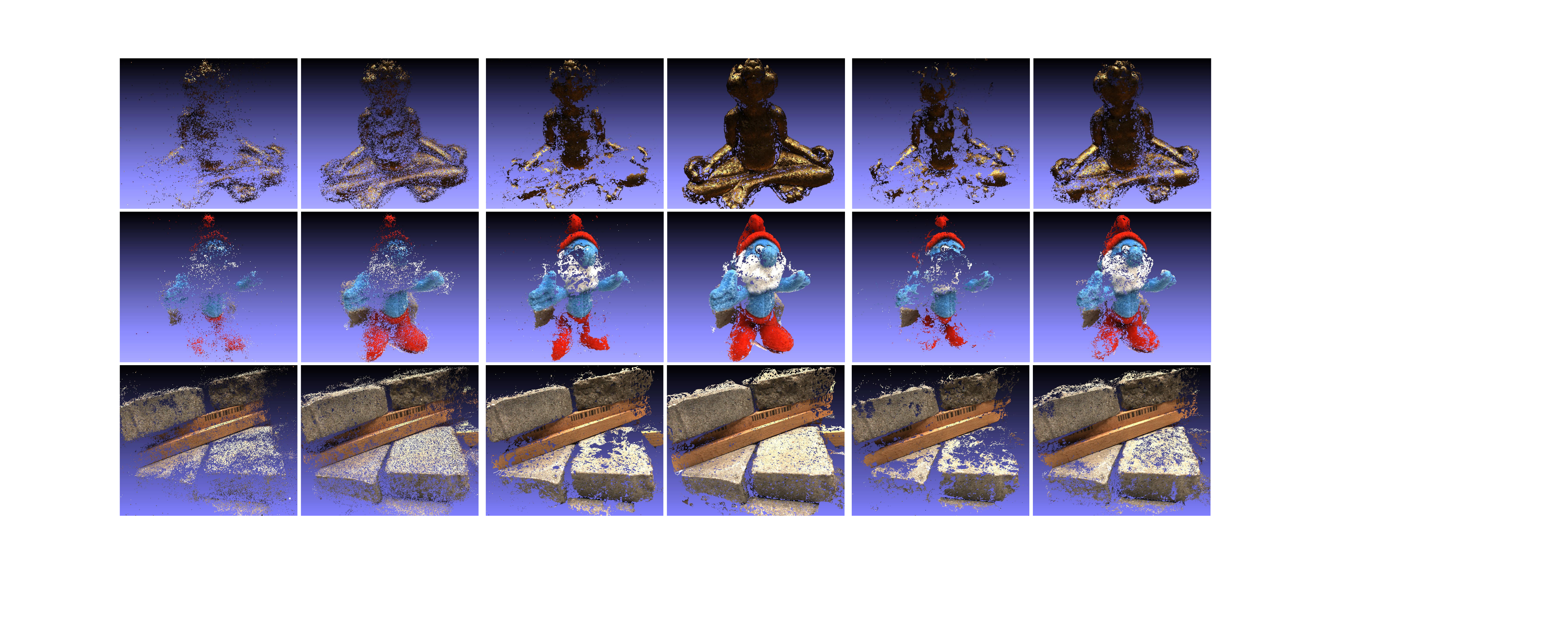}
\leftline{\hspace{0.005\textwidth} TransMVSNet \cite{ding2022transmvsnet} \hspace{0.03\textwidth} \cite{ding2022transmvsnet}+Ours \hspace{0.045\textwidth}  CasMVSNet \cite{gu2020cascade} \hspace{0.04\textwidth}  \cite{gu2020cascade}+Ours \hspace{0.07\textwidth}  UCSNet \cite{cheng2020deep} \hspace{0.06\textwidth}  \cite{cheng2020deep}+Ours}
\end{center}
\caption{Additional point cloud visualization on DTU. Results improve in all combinations of our method with different MVS models.}
\label{fig: dtu_pcd_sup}
\end{figure}

\begin{figure}[htb]
\begin{center}
\rotatebox{90}{\hspace{0.05\textwidth} \cite{gu2020cascade} + Ours \hspace{0.072\textwidth}  CasMVSNet \cite{gu2020cascade}}
\includegraphics[width=0.95\textwidth]{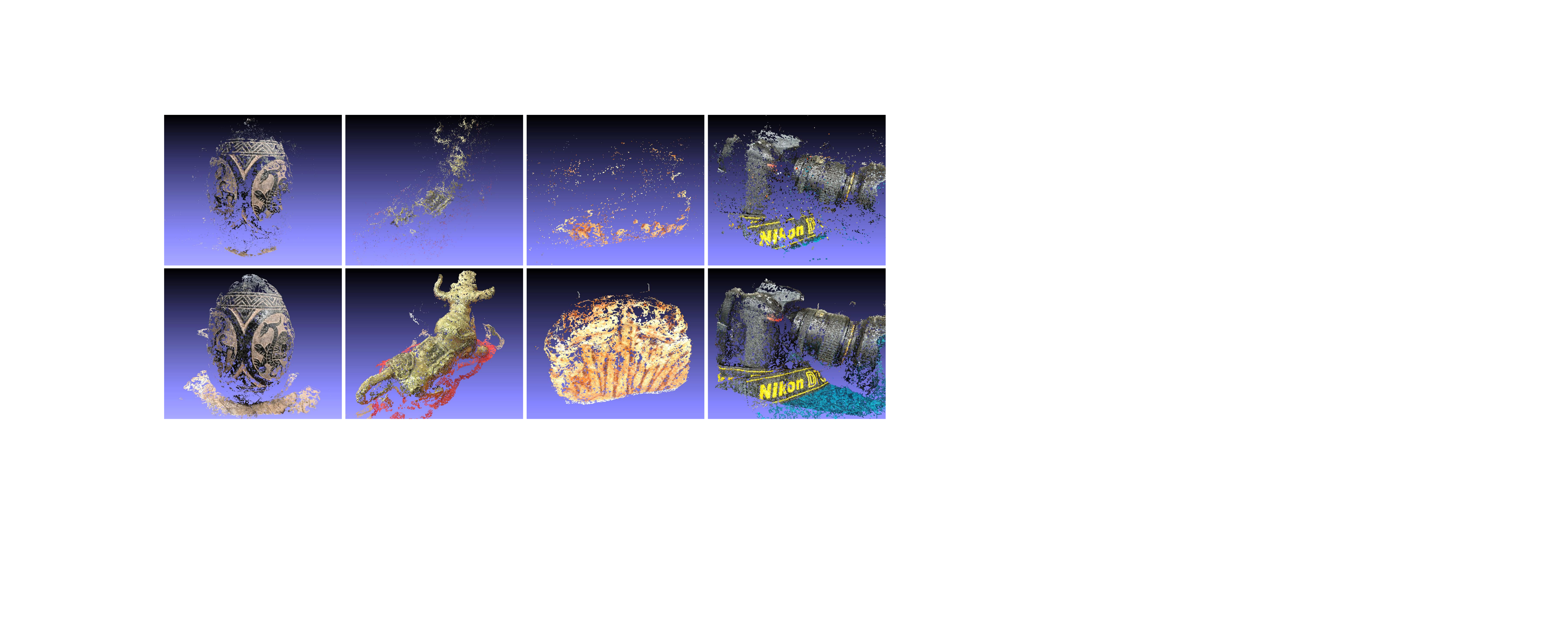}
\end{center}
\caption{Point cloud visualization on BlendedMVS when combining our method with CasMVSNet \cite{gu2020cascade}.}
\label{fig: bmvs_pcd_sup}
\end{figure}

\noindent \textbf{Additional Results on 3D Reconstructions.} We showcase additional meshes extracted from neural rendering methods on three-view 3D reconstruction for the DTU \cite{aanaes2016large} and BlendedMVS \cite{yao2020blendedmvs} datasets (\cref{fig: dtu_mesh_sup} and \cref{fig: bmvs_mesh_sup}). We provide more point cloud visualizations of the results when combining our method with different MVS models in \cref{fig: dtu_pcd_sup} and \cref{fig: bmvs_pcd_sup}.

\begin{figure}[!htb]
\begin{center}
\includegraphics[width=1.0\textwidth]{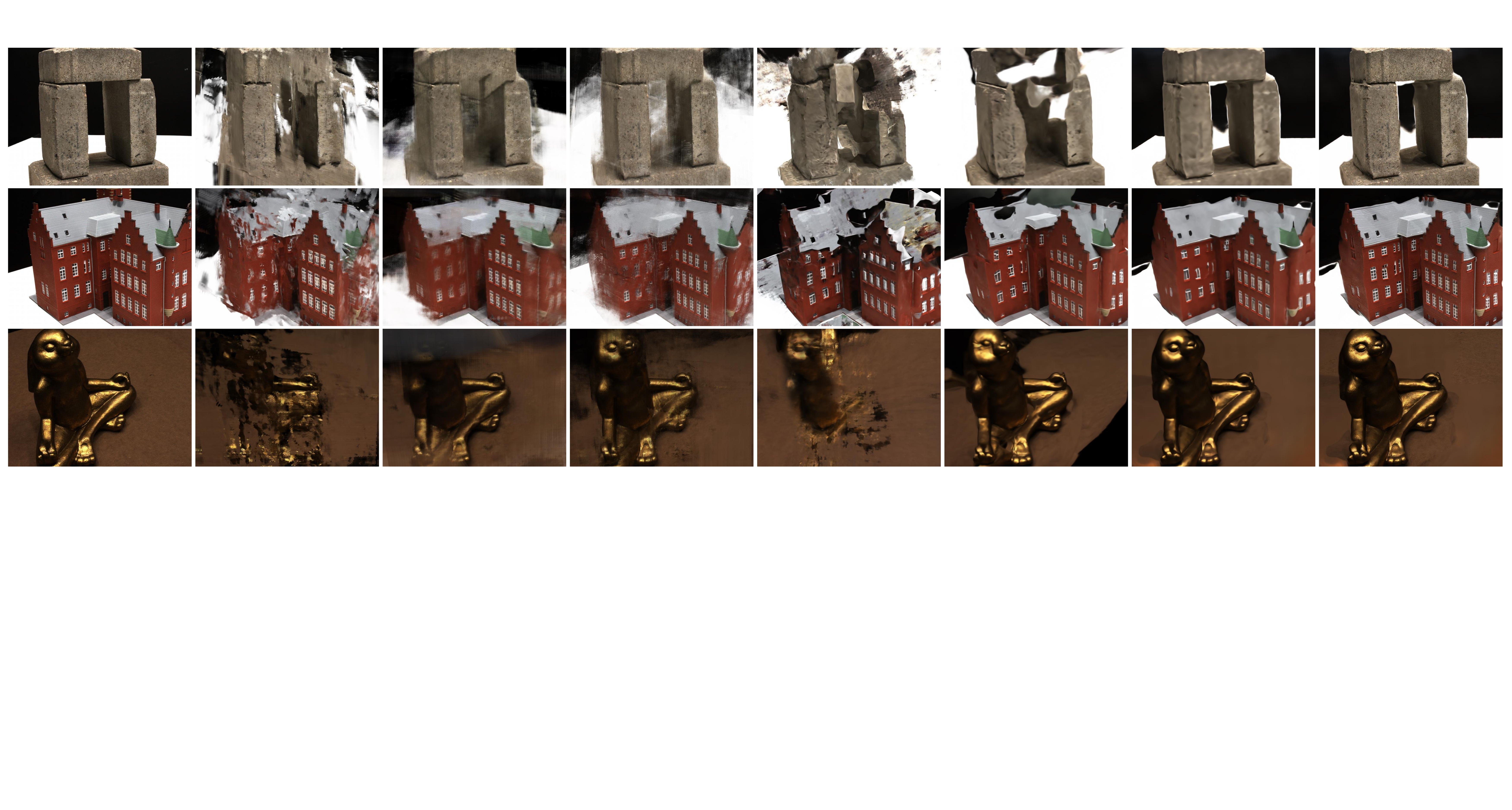}
\leftline{ \hspace{0.04\textwidth}    GT    \hspace{0.04\textwidth}  IBRNet$_{ft}$ \cite{wang2021ibrnet} \hspace{0.0\textwidth} MVSNeRF \cite{chen2021mvsnerf} \hspace{0.0\textwidth} GeoNeRF \cite{johari2022geonerf} \hspace{0.01\textwidth}  NeuS \cite{wang2021neus}    \hspace{0.015\textwidth} VolSDF \cite{yariv2021volume}  \hspace{0.04\textwidth}    \textbf{Ours}            \hspace{0.065\textwidth}     \textbf{Ours$_{IR}$}}
\end{center}
\caption{Additional novel view synthesis comparison on DTU. Our method leads to more accurate novel views.}
\label{fig: dtu_nvs_sup}
\end{figure}

\noindent \textbf{Additional Results on Novel View Synthesis.} In \cref{fig: dtu_nvs_sup} and \cref{fig: bmvs_mesh_sup} we showcase additional qualitative comparisons between our method and the baselines on novel view synthesis for the DTU and BlendedMVS datasets.

\begin{figure}[htb]
\begin{center}
\includegraphics[width=0.7\textwidth]{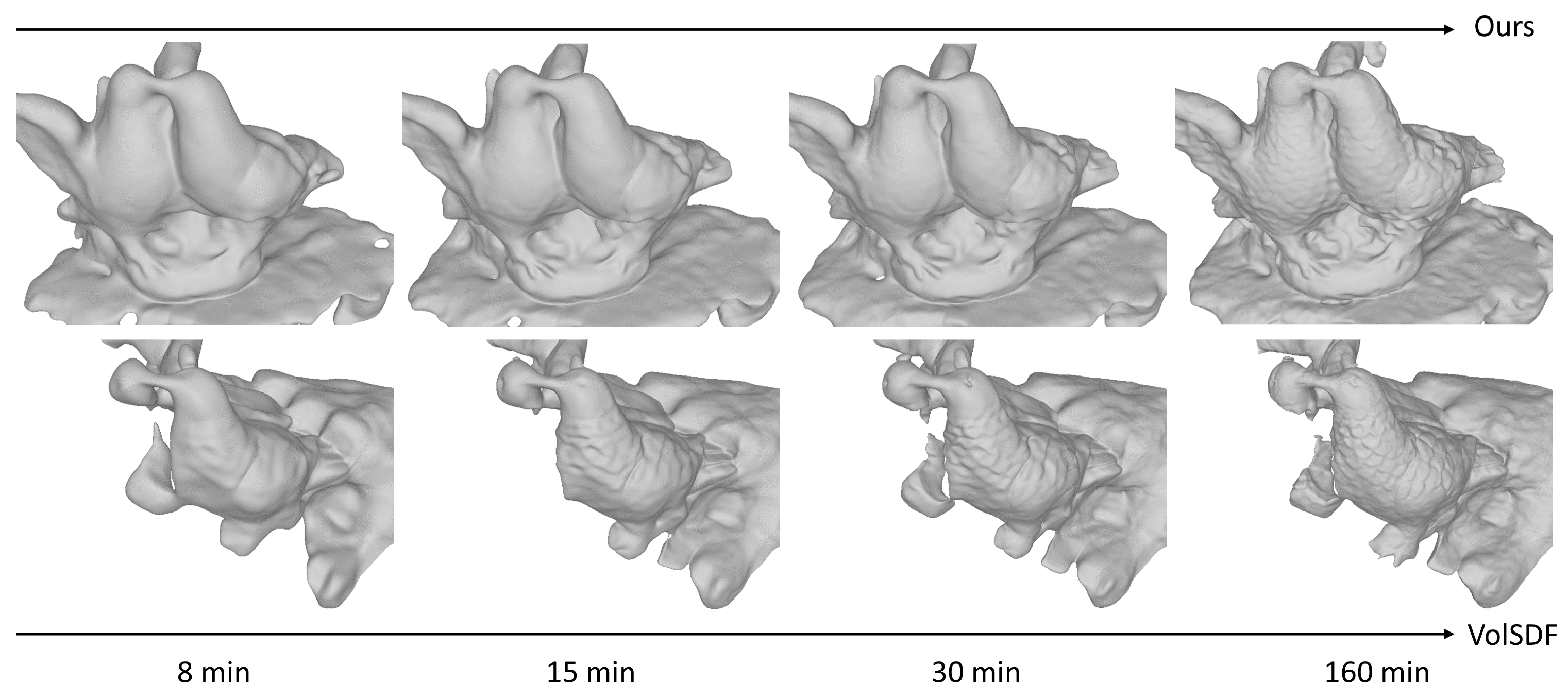}
\end{center}
\caption{An example of the implicit surface during the optimization process. We show that, with only 10-15 minutes of training, our output surface reconstruction is already reasonably good to guide finer stage of MVS, compared to the sub-optimal results of VolSDF \cite{yariv2021volume}.}
\label{fig: speed_sup}
\end{figure}

\noindent \textbf{Optimization Process.} In \cref{fig: speed_sup}, we show an example of how the implicit surface evolves during the optimization process. Our output surface reconstruction after 10-15 minutes of training (on an NVIDIA A5000 GPU) is already more accurate than the reconstruction of a fully trained VolSDF \cite{yariv2021volume} (typically 4-10 hours).

\begin{figure}[!htb]
\centering
\includegraphics[width=0.4\linewidth]{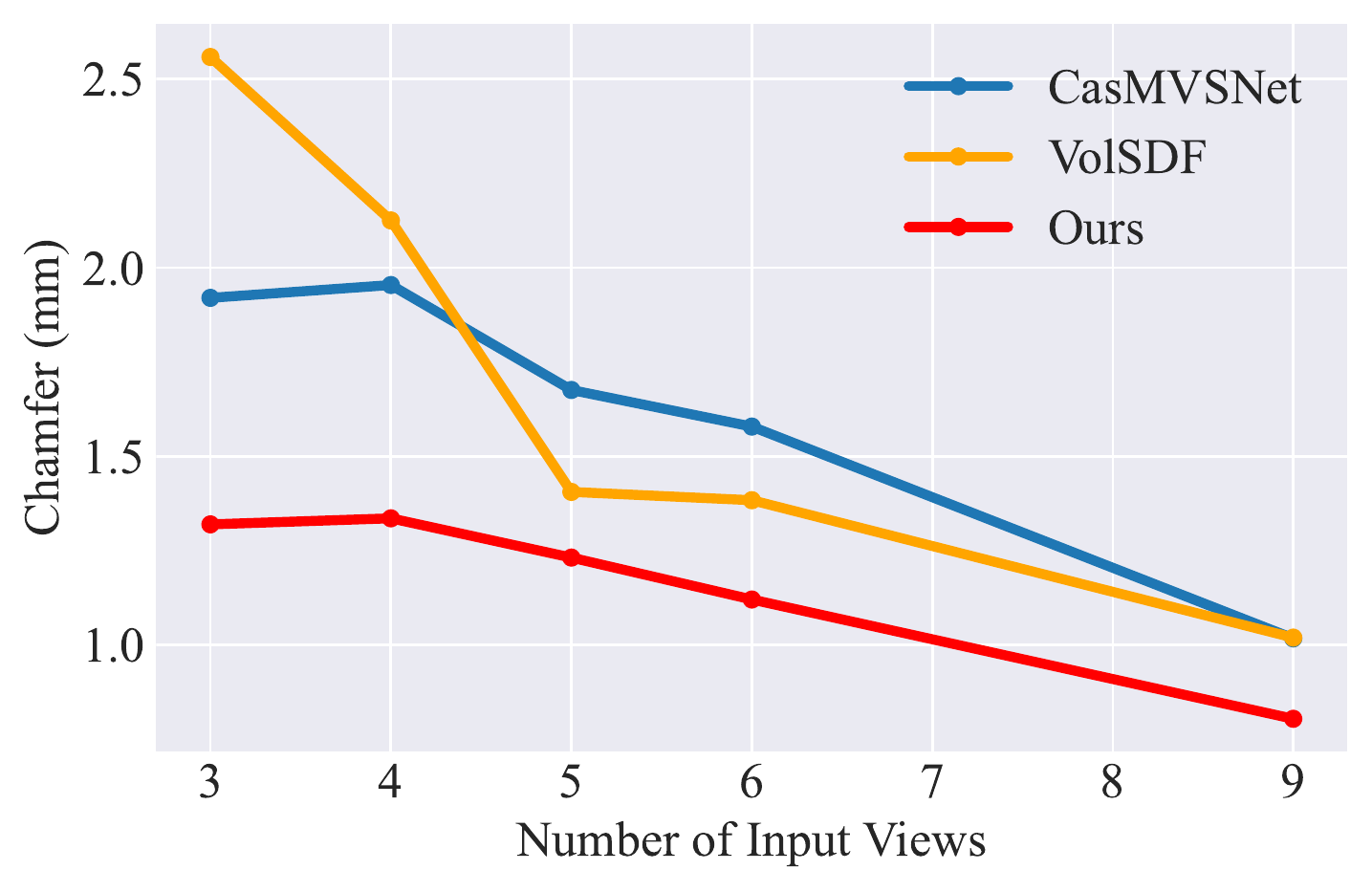}
\caption{Quantitative results on 3D reconstruction with 3-9 input views on DTU.}
\label{fig: n_views}
\end{figure}

\noindent \textbf{Scalability.} We conduct an ablation study on the scalability of our method. \cref{fig: n_views} and \cref{tab: n_views_nvs} show that as the input views become denser, the performance of our method, measured by surface reconstruction and novel view synthesis quality, improves and is consistently better than CasMVSNet \cite{gu2020cascade} and VolSDF \cite{yariv2021volume}. \cref{tab: higher_res} shows that, for three given views, the reconstruction quality of our method remains the same when varying the input image resolution. CasMVSNet \cite{gu2020cascade} and VolSDF \cite{yariv2021volume} perform worse when lowering the image resolution.

\begin{table}[!htb]
\centering
\begin{tabular}{l|ccc|ccc|ccc}
\toprule
 & \multicolumn{3}{c}{PSNR $\uparrow$} & \multicolumn{3}{c}{SSIM $\uparrow$} & \multicolumn{3}{c}{LPIPS $\downarrow$} \\
 & 3-views & 6-views & 9-views & 3-views & 6-views & 9-views & 3-views & 6-views & 9-views \\
 \midrule
VolSDF \cite{yariv2021volume} & 16.99 & 20.19 & \textbf{23.04} & 0.786 & 0.823 & 0.836 & 0.332 & 0.317 & 0.310 \\
\textbf{Ours} & 20.21 & 20.80 & 22.98 & 0.820 & 0.824 & 0.832 & 0.321 & 0.318 & 0.309 \\
\textbf{Ours$_{IR}$} & \textbf{20.58} & \textbf{21.48} & 23.01 & \textbf{0.855} & \textbf{0.872} & \textbf{0.895} & \textbf{0.157} & \textbf{0.145} & \textbf{0.128} \\
\bottomrule
\end{tabular}
\caption{Quantitative results on novel view synthesis with 3-9 input views on DTU.}
\label{tab: n_views_nvs}
\end{table}

\begin{table}[!htb]
\centering
\begin{tabular}{l|ccc|ccc|ccc|ccc}
\toprule
 & \multicolumn{3}{c}{Chamfer $\downarrow$} & \multicolumn{3}{c}{PSNR $\uparrow$} & \multicolumn{3}{c}{SSIM $\uparrow$} & \multicolumn{3}{c}{LPIPS $\downarrow$} \\
 Resolution & Low & Mid & High & Low & Mid & High & Low & Mid & High & Low & Mid & High \\
\midrule
CasMVSNet \cite{gu2020cascade} & 1.92 & 1.86 & 1.87 & \multicolumn{3}{c}{---} & \multicolumn{3}{c}{---} & \multicolumn{3}{c}{---}  \\
VolSDF \cite{yariv2021volume} & 2.56 & 2.80 & 2.70 & 16.99 & 15.52 & 15.75 & 0.786 & 0.771 & 0.790 & 0.332 & 0.352 & 0.346 \\
\textbf{Ours} & \textbf{1.32} & \textbf{1.33} & \textbf{1.33} & 20.21 & 19.63 & 19.97 & 0.820 & 0.822 & 0.833 & 0.321 & 0.330 & 0.330 \\
\textbf{Ours$_{IR}$} & \multicolumn{3}{c}{---} & \textbf{20.58} & \textbf{19.98} & \textbf{20.30} & \textbf{0.855} & \textbf{0.853} & \textbf{0.858} & \textbf{0.157} & \textbf{0.178} & \textbf{0.186} \\
\bottomrule
\end{tabular}
\caption{Quantitative results with different image resolutions: low (576$\times$768), mid (864$\times$1152), and high (1152$\times$1536), on DTU.}
\label{tab: higher_res}
\end{table}



\noindent \textbf{Ablation Study on Different MVS Models.} In \cref{tab: mvsbackbones}, we provide an extended ablation study across all three MVS models: TransMVSNet \cite{ding2022transmvsnet}, UCSNet \cite{cheng2020deep}, and CasMVSNet \cite{gu2020cascade}. It validates the importance of using probability volumes, soft consistency check, and generalized cross-entropy loss, consistent with our main text’s ablation study findings.

\begin{table}[!htb]
\centering
\begin{tabular}{lccc}
\toprule
Chamfer (mm)$\downarrow$ & TransMVSNet \cite{ding2022transmvsnet} & UCSNet \cite{cheng2020deep} & CasMVSNet \cite{gu2020cascade} \\
\midrule
MVS Model & 2.915 & 2.201 & 1.920 \\
MVS + \textbf{Ours} & 1.798 & 1.519 & \textbf{1.320} \\
only soft consistency & 2.627 & 1.901 & 1.711 \\
MSE loss & 2.233 & 2.019 & 1.792 \\
w/o prob. volume & 2.692 & 1.791 & 1.543 \\
w/o GCE loss & 2.525 & 1.702 & 1.534 \\
\bottomrule
\end{tabular}
\caption{Ablation study on different MVS models, on DTU.}
\label{tab: mvsbackbones}
\end{table}
 
\noindent \textbf{Additional Comparison with Related Work.} In \cref{tab: related_work}, we provide additional comparisons with regularization based approach including DS-NeRF \cite{deng2022depth}, which utilizes estimated depth from structure-from-motion \cite{schonberger2016structure}, and MonoSDF \cite{yu2022monosdf}, which \cite{wang2023sparsenerf} utilizes monocular depth estimation. Because their depth priors are either sparse or often not accurate enough, providing only approximated structures or shapes, their results are worse than ours.

\begin{table}[!htb]
\centering
\begin{tabular}{lccc}
\toprule
& MonoSDF \cite{yu2022monosdf} & DS-NeRF \cite{deng2022depth} & \textbf{Ours} \\
\midrule
Chamfer (mm)$\downarrow$ & 2.141 & 1.792 & \textbf{1.32} \\
\bottomrule
\end{tabular}
\caption{Additional comparison with related work, on DTU.}
\label{tab: related_work}
\end{table}

\noindent \textbf{Limitations.} While our method is also capable of refining the probability volumes of the finer stages of MVS, we notice that the benefits diminsh since there is not as much uncertainty in later stages. Our method applied to stages 1, 1,2, and 1,2,3 of MVS resulted in chamfer distances of 1.320, 1.312, and 1.309, respectively.

\begin{figure}[!htb]
    \centering
    \includegraphics[width=0.8\linewidth]{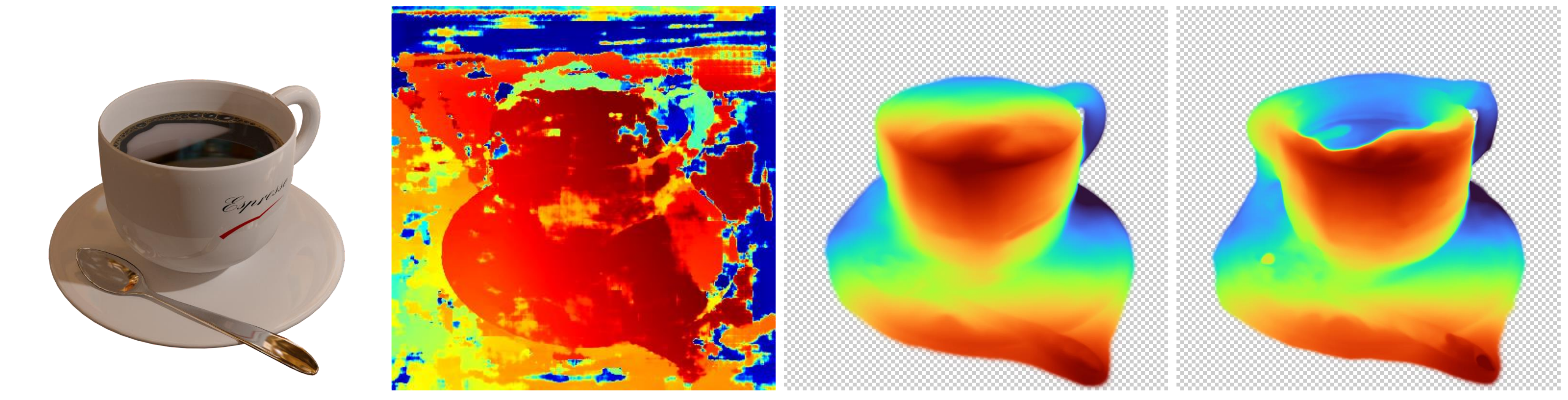}
    \leftline{
    \hspace{0.18\textwidth} GT \hspace{0.11\textwidth}  CasMVSNet \cite{gu2020cascade} \hspace{0.07\textwidth}  VolSDF \cite{yariv2021volume}
    \hspace{0.11\textwidth}  Ours}
    \caption{Depth map predictions on Shiny Dataset.}
    \label{fig: shiny_compare}
\end{figure}

\noindent \textbf{Evaluation on Objects with Glossy Material.} Although our method may not work well for texture-less or glossy surfaces due to the introduction of MVS. Surprisingly, as shown in \cref{fig: shiny_compare} and \cref{tab: shiny}, our method still surpasses VolSDF in reconstructing complex glossy surfaces. We suspect that our noise-tolerant optimization and MVS models operating on features instead of pixels make our pipeline more robust to specular reflections that violate multi-view consistency. Further research on this problem would be quite interesting.

\begin{table}[!htb]
\centering
\begin{tabular}{lcccc}
\toprule
& PSNR $\uparrow$ & SSIM $\uparrow$ & LPIPS $\downarrow$ & MAE$^\circ$ $\downarrow$ \\
\midrule
VolSDF \cite{yariv2021volume} & 20.71 & 0.943 & 0.126 & 32.96 \\
\textbf{Ours} & 20.97 & 0.944 & 0.124 & \textbf{29.26} \\
\textbf{Ours$_{IR}$} & \textbf{21.50} & \textbf{0.944} & \textbf{0.081} & \\
\bottomrule
\end{tabular}
\caption{Results on Shiny Dataset (6 scenes, from Ref-NeRF \cite{verbin2022ref}). Mean angular error (MAE) is used in evaluating normal vectors.}
\label{tab: shiny}
\end{table}

\section{Experimental Settings}
\label{sec: add_impl}

\noindent \textbf{Hyperparameters.} We observe a strong over-fitting tendency for VolSDF \cite{yariv2021volume} with sparse input views. This over-fitting is due to the usage of the view direction to explain object color in different views, and therefore we set the positional encoding level of view direction to 1 for VolSDF and our method. We use the same loss functions as VolSDF \cite{yariv2021volume}, along with our weight loss $\mathcal{L}_{weight}$ and a sparsity regularization $\mathcal{L}_{sparse}$. Both $\mathcal{L}_{weight}$ and $\mathcal{L}_{sparse}$ are weighted with a value of 1.0. The $\epsilon$ in $\mathcal{L}_{sparse}$ is 0.001. Moreover, we do not apply weight loss for rays with weak MVS supervision (i.e. the sum of consistency-weighted probability along the ray is less than 0.001). We found that our weight loss is highly tolerant to parameter choices. We used grid search to find the best $q$ but determined that all $q$ in $[0.2,0.8]$ yield satisfactory results (overall error: 1.32-1.44). We set $q=0.5$ in all our experiments. 

\noindent \textbf{Rendering Pipeline.} In testing, our method utilizes image-based rendering. We merge source pixels from multiple source images for a target pixel. More specifically, we first render depth maps for all source views. Then, for a target view, we render its depth map and project its pixels back to the source views and we apply consistency check on the back-projected depths with the source depth to determine its visibility on source views and retrieve the interpolated source pixel colors. The blending weights for pixel colors from different source views are based on the cosine between the target and source pixels' view directions, computed using \emph{softmax} with a temperature of 20. In areas where there are no valid pixels to blend (i.e., the geometric consistency check fails for all source views), we use the rendered colors. Finally, a 4-level Laplacian pyramid \cite{burt1987laplacian} is used to smoothly blend source pixels.

\noindent \textbf{MVS Models.} In our experiments, we compare our proposed method against TransMVSNet \cite{ding2022transmvsnet}, CasMVSNet \cite{gu2020cascade}, and UCSNet \cite{cheng2020deep}. We employ the official implementation of each method provided by the authors and use their published pre-trained models. To ensure a fair comparison, the weights for all three models we used were pre-trained exclusively on the DTU dataset \cite{aanaes2016large} with ground-truth depth as supervision.

\noindent \textbf{Denser Plane Sweep.} The main difference in our training scheme, compared to MVS models, is the usage of a denser plane sweep, which we also implemented for all baseline MVS models, reducing their overall error by 33\% on average.

\noindent \textbf{The Choice of CasMVSNet and VolSDF.}  In our method, we select CasMVSNet \cite{gu2020cascade} as the MVS model and VolSDF \cite{yariv2021volume} as the neural rendering model. We opt for CasMVSNet as it is the representative coarse-to-fine MVS model, and we find no substantial improvement in other recent MVS models when compared to CasMVSNet for sparse-input scenarios, as demonstrated in the main text. We use VolSDF, which is a state-of-the-art implicit surface reconstruction method, as demonstrated in \cite{long2022sparseneus, yariv2021volume}. Nevertheless, other neural rendering models like NeRF \cite{mildenhall2020nerf} and NeuS \cite{wang2021neus} can also be used in our method but the differences in the overall performance are a subject for future work.

\begin{figure}[!htb]
\centering
\includegraphics[width=0.4\linewidth]{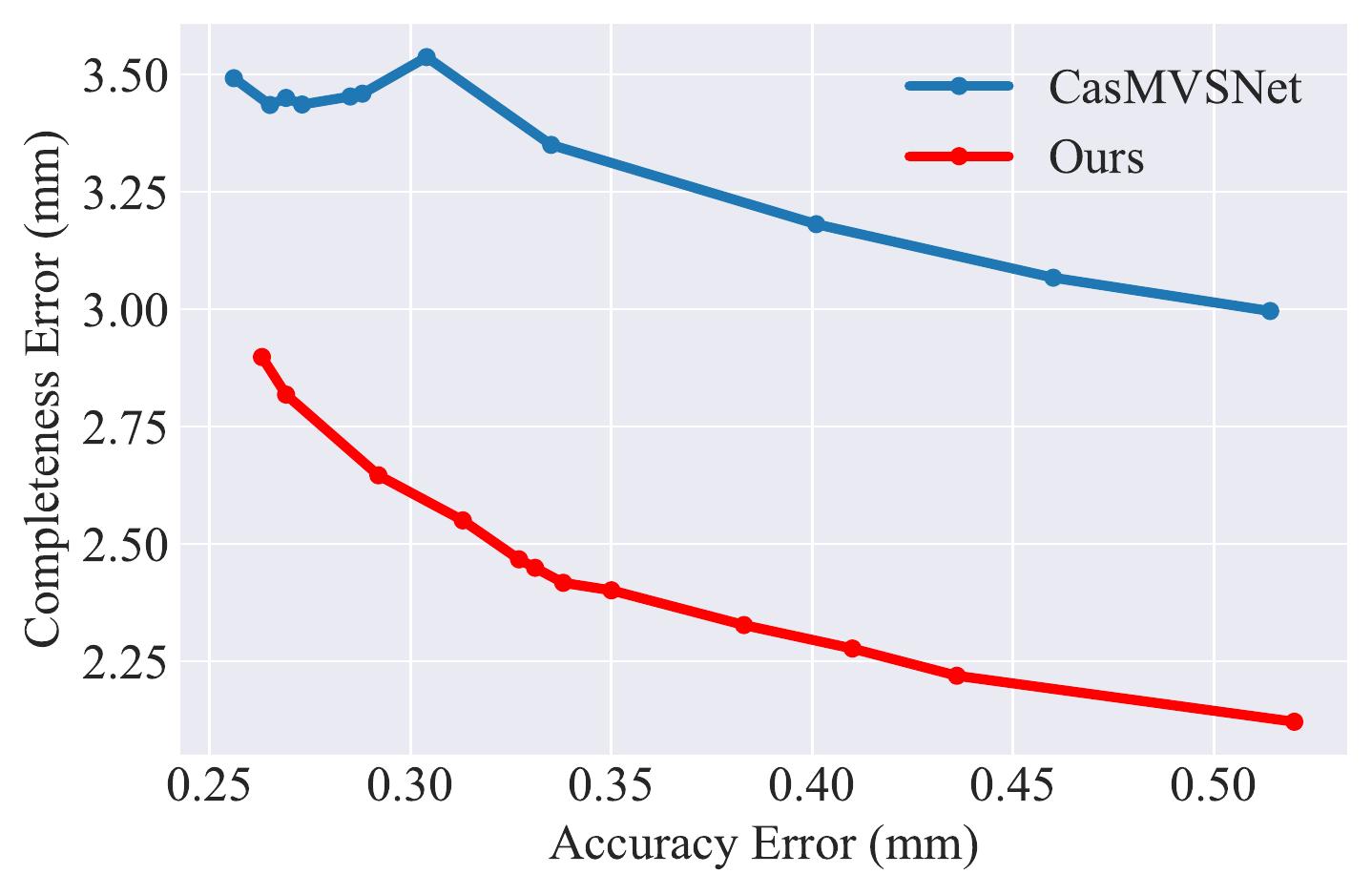}
\caption{Completeness error and Accuracy error trade-off.}
\label{fig: acc_comp_sup}
\end{figure}

\noindent \textbf{Metrics.} The Chamfer distance is the average of the Accuracy (the distance from the reconstructed point cloud to reference) and Completeness (the distance from reference to reconstruction). The use of stronger geometric/photometric filtering can lead to better accuracy, but at the expense of completeness, and vice-versa. Given this trade-off between accuracy and completeness in point cloud filtering, we choose to employ the Chamfer distance metric as our primary measure in the main text, following \cite{yariv2020multiview, yariv2021volume}. We present the Accuracy-Completeness trade-off in \cref{fig: acc_comp_sup}. The results reveal that we consistently attain roughly 30\% higher completeness than the baseline across all accuracy levels.

\noindent \textbf{Datasets.} For the DTU dataset \cite{aanaes2016large}, we combine the scans used in \cite{yariv2020multiview, yariv2021volume, yu2021pixelnerf} with the ones used in conventional MVS settings \cite{ding2022transmvsnet, yao2018mvsnet}, and remove the training scans of common MVS models. Specifically, we use scans 21, 24, 34, 37, 38, 40, 82, 106, 110, 114, and 118 for our evaluation. For evaluation on DTU, we adhere to the standard protocol in \cite{aanaes2016large, yariv2021volume, niemeyer2022regnerf}.

\setlength{\parskip}{0mm}

The BlendedMVS dataset \cite{yao2020blendedmvs} lacks a standard evaluation protocol for sparse-view scenarios. Therefore, we adopted a similar evaluation protocol to DTU; select three sparse input views with a relatively wide baseline and evaluate using object masks. Similar to DTU, only scene objects are used in the evaluation. This is simply performed by removing the plane from the ground truth point cloud. The sparse view indexes we adopt are: Doll: 9, 10, 55; Egg: 9, 52, 59; Head: 22, 26, 27; Angel: 11, 39, 53; Bull: 32, 42, 47; Robot: 28, 34, 57; Dog: 2, 5, 25; Bread: 16, 21, 33; Camera: 10, 16, 60. For reference, we offer quantitative comparisons without using object masks or removing the plane in \cref{tab: bmvs_nomask}.

\begin{table}[!htb]\centering
\begin{tabular}{lcccccccccc}
\toprule
Scene & Doll & Egg & Head & Angel & Bull & Robot & Dog & Bread & Camera & \textbf{Mean} \\
\midrule
MVSNeRF \cite{chen2021mvsnerf} & 22.3 & -9.7 & -30.8 & 38.1 & 4.1 & 24.8 & -2.7 & 2.7 & 8.6 & 6.4 \\
GeoNeRF \cite{johari2022geonerf} & \textbf{48.8} & 37.9 & 3.6 & 37.6 & -7.8 & 30.3 & 29.4 & 19.1 & 9.2 & 23.1 \\
\cmidrule(l{0.7em}r{0.7em}){1-11}
CasMVSNet  \cite{gu2020cascade} & 46.2 & 47.7 & -0.2 & 45.8 & -6.6 & 41.5 & 41.3 & 8.9 & 31.8 & 28.5 \\
\textbf{Ours} & 47.8 & \textbf{62.0} & \textbf{23.3} & \textbf{54.7} & \textbf{20.6} & \textbf{49.7} & \textbf{48.0} & \textbf{59.9} & \textbf{49.3} & \textbf{46.1} \\
\bottomrule
\end{tabular}
\caption{BlendedMVS 3D reconstruction results without applying object masks on the reconstruction results. Since there are no units in BlendedMVS, we report relative improvement (in \%) over VolSDF \cite{yariv2021volume} in terms of Chamfer distance.}
\label{tab: bmvs_nomask}
\end{table}

In the context of novel view synthesis, it is noteworthy that while the BlendedMVS dataset has 360-degree views of an object, the sparse inputs partially cover the frontal area. Consequently, conducting novel view synthesis on all images, including the back views, is unreasonable. Therefore, we choose to evaluate the closest 12 views in each scene. The indexes for evaluation are: Doll: 0, 13, 19, 20, 22, 31, 33, 35, 36, 37, 58, 61; Egg: 1, 8, 12, 14, 23, 27, 37, 39, 49, 65, 68, 71; Head: 0, 1, 6, 7, 11, 13, 15, 16, 25, 28, 31, 33; Angel: 0, 2, 9, 23, 29, 30, 46, 48, 50, 59, 68, 71; Bull: 0, 13, 16, 17, 20, 24, 26, 41, 44, 55, 57, 58; Robot: 1, 2, 10, 13, 22, 25, 40, 55, 73, 75, 80, 88; Dog: 0, 6, 7, 8, 10, 13, 14, 17, 22, 23, 27, 29; Bread: 8, 10, 17, 18, 24, 25, 26, 27, 28, 30, 43, 47; Camera: 18, 25, 59, 65, 68, 83, 89, 92, 94, 118, 133, 136.




\end{document}